\documentclass[10pt,twocolumn,letterpaper]{article}
\usepackage[pagenumbers]{cvpr}

\usepackage{cuted} 

\usepackage{currfile} 

\usepackage{caption} 

\usepackage{amsmath,amssymb,booktabs,graphicx,colortbl,adjustbox,multirow}
\usepackage{accsupp}
\definecolor{benchmarkgroup}{HTML}{EFEFEF}
\definecolor{phoenixrow}{HTML}{F5E9DD}
\definecolor{cvprblue}{rgb}{0.21,0.49,0.74}
\usepackage[breaklinks,colorlinks,allcolors=cvprblue]{hyperref}
\hypersetup{
  pdftitle={Phoenix: Learning Scene Structure Representations for Generalizable World-Action Models},
  pdfauthor={Yupeng Zheng, Chaoyue Li, Linbo Wang, Xiang Li, Yuhang Zheng, Xiaotong Su, Songen Gu, Yichen Zhang, Zebin Xing, Shuai Tian, Jingxiang Guo, Jing Li, Zhiyong Zhao, Zhongpu Xia, Qichao Zhang, Haoran Li, Shuicheng Yan, Dongbin Zhao},
  pdfsubject={World-action modeling, scene representations, and robotic manipulation}
}
\def\confName{CVPR}
\def\confYear{2027}
\newcommand{\PhoenixLogo}{%
  \raisebox{-5pt}{\includegraphics[trim=35bp 118bp 1620bp 105bp,clip,height=24pt]{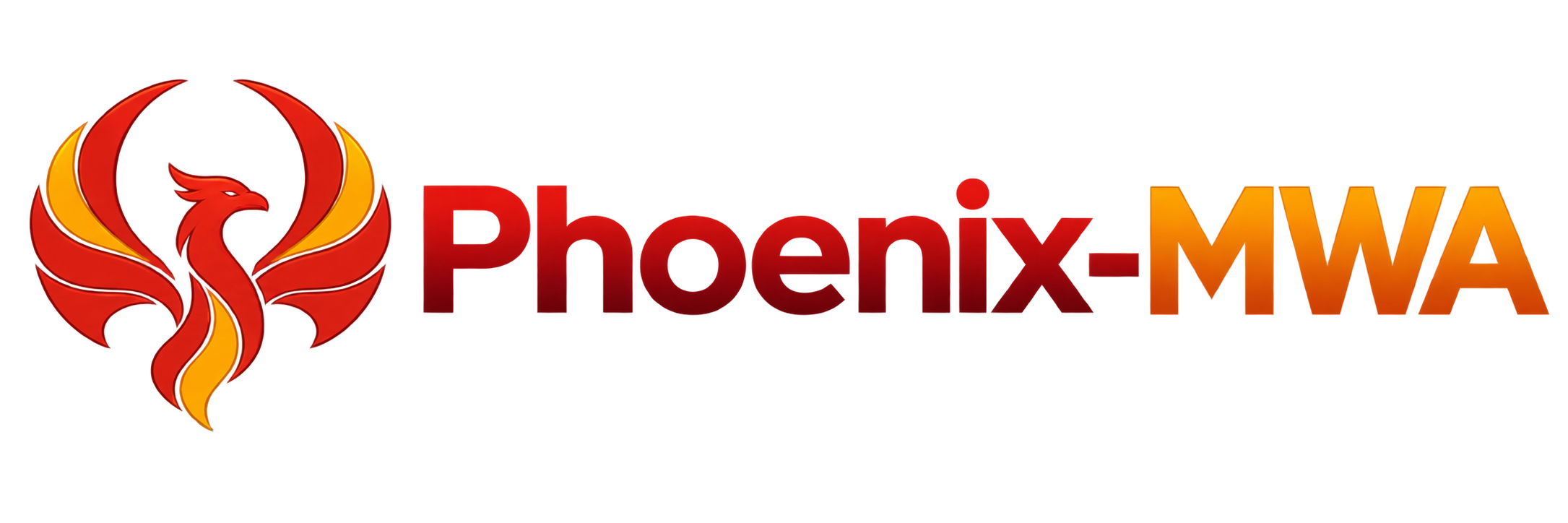}}%
}
\newcommand{\PhoenixTitleWord}{%
  \BeginAccSupp{method=escape,ActualText={Phoenix}}%
  \raisebox{-0.25pt}{\includegraphics[trim=580bp 282bp 663bp 242bp,clip,height=15pt]{images/phoenix-brand-original.png}}%
  \EndAccSupp{}%
}
\usepackage{tikz}
\usetikzlibrary{svg.path}

\newcommand{\PhoenixPageIcon}{%
  \raisebox{-1pt}{\resizebox{!}{9pt}{%
    \begin{tikzpicture}[x=1pt,y=1pt,yscale=-1,baseline=(current bounding box.south)]
      \path[use as bounding box] (0,0) rectangle (512,512);
      \fill svg {M352 256c0 22.2-1.2 43.6-3.3 64l-185.3 0c-2.2-20.4-3.3-41.8-3.3-64s1.2-43.6 3.3-64l185.3 0c2.2 20.4 3.3 41.8 3.3 64zm28.8-64l123.1 0c5.3 20.5 8.1 41.9 8.1 64s-2.8 43.5-8.1 64l-123.1 0c2.1-20.6 3.2-42 3.2-64s-1.1-43.4-3.2-64zm112.6-32l-116.7 0c-10-63.9-29.8-117.4-55.3-151.6c78.3 20.7 142 77.5 171.9 151.6zm-149.1 0l-176.6 0c6.1-36.4 15.5-68.6 27-94.7c10.5-23.6 22.2-40.7 33.5-51.5C239.4 3.2 248.7 0 256 0s16.6 3.2 27.8 13.8c11.3 10.8 23 27.9 33.5 51.5c11.6 26 20.9 58.2 27 94.7zm-209 0L18.6 160C48.6 85.9 112.2 29.1 190.6 8.4C165.1 42.6 145.3 96.1 135.3 160zM8.1 192l123.1 0c-2.1 20.6-3.2 42-3.2 64s1.1 43.4 3.2 64L8.1 320C2.8 299.5 0 278.1 0 256s2.8-43.5 8.1-64zM194.7 446.6c-11.6-26-20.9-58.2-27-94.6l176.6 0c-6.1 36.4-15.5 68.6-27 94.6c-10.5 23.6-22.2 40.7-33.5 51.5C272.6 508.8 263.3 512 256 512s-16.6-3.2-27.8-13.8c-11.3-10.8-23-27.9-33.5-51.5zM135.3 352c10 63.9 29.8 117.4 55.3 151.6C112.2 482.9 48.6 426.1 18.6 352l116.7 0zm358.1 0c-30 74.1-93.6 130.9-171.9 151.6c25.5-34.2 45.2-87.7 55.3-151.6l116.7 0z};
    \end{tikzpicture}}}%
}

\newcommand{\PhoenixCodeIcon}{%
  \raisebox{-1pt}{\resizebox{!}{9pt}{%
    \begin{tikzpicture}[x=1pt,y=1pt,yscale=-1,baseline=(current bounding box.south)]
      \path[use as bounding box] (0,0) rectangle (496,512);
      \fill svg {M165.9 397.4c0 2-2.3 3.6-5.2 3.6-3.3.3-5.6-1.3-5.6-3.6 0-2 2.3-3.6 5.2-3.6 3-.3 5.6 1.3 5.6 3.6zm-31.1-4.5c-.7 2 1.3 4.3 4.3 4.9 2.6 1 5.6 0 6.2-2s-1.3-4.3-4.3-5.2c-2.6-.7-5.5.3-6.2 2.3zm44.2-1.7c-2.9.7-4.9 2.6-4.6 4.9.3 2 2.9 3.3 5.9 2.6 2.9-.7 4.9-2.6 4.6-4.6-.3-1.9-3-3.2-5.9-2.9zM244.8 8C106.1 8 0 113.3 0 252c0 110.9 69.8 205.8 169.5 239.2 12.8 2.3 17.3-5.6 17.3-12.1 0-6.2-.3-40.4-.3-61.4 0 0-70 15-84.7-29.8 0 0-11.4-29.1-27.8-36.6 0 0-22.9-15.7 1.6-15.4 0 0 24.9 2 38.6 25.8 21.9 38.6 58.6 27.5 72.9 20.9 2.3-16 8.8-27.1 16-33.7-55.9-6.2-112.3-14.3-112.3-110.5 0-27.5 7.6-41.3 23.6-58.9-2.6-6.5-11.1-33.3 2.6-67.9 20.9-6.5 69 27 69 27 20-5.6 41.5-8.5 62.8-8.5s42.8 2.9 62.8 8.5c0 0 48.1-33.6 69-27 13.7 34.7 5.2 61.4 2.6 67.9 16 17.7 25.8 31.5 25.8 58.9 0 96.5-58.9 104.2-114.8 110.5 9.2 7.9 17 22.9 17 46.4 0 33.7-.3 75.4-.3 83.6 0 6.5 4.6 14.4 17.3 12.1C428.2 457.8 496 362.9 496 252 496 113.3 383.5 8 244.8 8zM97.2 352.9c-1.3 1-1 3.3.7 5.2 1.6 1.6 3.9 2.3 5.2 1 1.3-1 1-3.3-.7-5.2-1.6-1.6-3.9-2.3-5.2-1zm-10.8-8.1c-.7 1.3.3 2.9 2.3 3.9 1.6 1 3.6.7 4.3-.7.7-1.3-.3-2.9-2.3-3.9-2-.6-3.6-.3-4.3.7zm32.4 35.6c-1.6 1.3-1 4.3 1.3 6.2 2.3 2.3 5.2 2.6 6.5 1 1.3-1.3.7-4.3-1.3-6.2-2.2-2.3-5.2-2.6-6.5-1zm-11.4-14.7c-1.6 1-1.6 3.6 0 5.9 1.6 2.3 4.3 3.3 5.6 2.3 1.6-1.3 1.6-3.9 0-6.2-1.4-2.3-4-3.3-5.6-2z};
    \end{tikzpicture}}}%
}

\title{\PhoenixLogo\hspace{0.30em}\PhoenixTitleWord:
  Learning Scene Structure Representations\\
  for Generalizable World-Action Models}
\newcommand{\PhoenixAuthor}[2]{\mbox{#1\textsuperscript{#2}}}
\author{%
  \PhoenixAuthor{Yupeng Zheng}{1,2,*,\textdagger}\quad
  \PhoenixAuthor{Chaoyue Li}{1,*}\quad
  \PhoenixAuthor{Linbo Wang}{1,*}\quad
  \PhoenixAuthor{Xiang Li}{3,*}\quad
  \PhoenixAuthor{Yuhang Zheng}{4}\quad
  \PhoenixAuthor{Xiaotong Su}{1}\\[2pt]
  \PhoenixAuthor{Songen Gu}{5}\quad
  \PhoenixAuthor{Yichen Zhang}{1}\quad
  \PhoenixAuthor{Zebin Xing}{1}\quad
  \PhoenixAuthor{Shuai Tian}{1}\quad
  \PhoenixAuthor{Jingxiang Guo}{4}\quad
  \PhoenixAuthor{Jing Li}{2}\\[2pt]
  \PhoenixAuthor{Zhiyong Zhao}{2}\quad
  \PhoenixAuthor{Zhongpu Xia}{2,\textdagger}\quad
  \PhoenixAuthor{Qichao Zhang}{1}\quad
  \PhoenixAuthor{Haoran Li}{1}\quad
  \PhoenixAuthor{Shuicheng Yan}{4}\quad
  \PhoenixAuthor{Dongbin Zhao}{1}\\[5pt]
  {\fontsize{9.2}{11.5}\selectfont
    \textsuperscript{1}Institute of Automation, Chinese Academy of Sciences\quad
    \textsuperscript{2}Anyverse Dynamics}\\[1pt]
  {\fontsize{9.2}{11.5}\selectfont
    \textsuperscript{3}Tsinghua University\quad
    \textsuperscript{4}National University of Singapore\quad
    \textsuperscript{5}Fudan University}\\[2pt]
  {\fontsize{8.5}{10}\selectfont \textsuperscript{*}Equal contribution.\quad \textsuperscript{\textdagger}Project leaders.}\\[0pt]
  {\fontsize{8.5}{10}\selectfont Contact: Yupeng Zheng (\href{mailto:zhengyupeng18@mails.ucas.ac.cn}{\texttt{zhengyupeng18@mails.ucas.ac.cn}})}\\[2pt]
  {\fontsize{10}{12}\selectfont
    \href{https://openphoenix-team.github.io/Phoenix-pages/}{\PhoenixPageIcon\hspace{0.35em}\textbf{Project page}}\quad$\vert$\quad
    \href{https://github.com/Anyverse-Dynamics/Phoenix-MWA}{\PhoenixCodeIcon\hspace{0.35em}\textbf{Code}}}
}

\makeatletter
\def\@maketitle{%
  \newpage\null\vskip .15in
  \begin{center}
    {\fontsize{17}{21}\selectfont\bfseries \@title\par}
    \vskip 8pt
    {\fontsize{10.5}{13.5}\selectfont \@author\par}
    \vskip 6pt
  \end{center}%
}
\makeatother
\begin{document}
\maketitle
\begin{abstract}
World-action models jointly predict future observations and actions, yet reconstruction-oriented visual representations do not explicitly emphasize the spatial relationships and object-state changes needed for manipulation.
We introduce Phoenix, which learns representations of scene structure for world-action modeling.
Specifically, Phoenix Encoder enriches DINOv3's general visual semantics through dual-teacher distillation of geometric and spatiotemporal semantic knowledge.
Next, CS-VAE causally aggregates and compresses Phoenix Encoder features into generative latents while preserving scene information and spatial layout.
Finally, Phoenix jointly predicts these latents, complementary appearance latents, and robot actions, linking scene evolution with action generation.
To study pretraining data efficiency, we curate approximately 39,000 hours of egocentric and robot manipulation data for representation learning and use the robot demonstration subset for WAM pretraining. 
Following benchmark-specific post-training, Phoenix achieves state-of-the-art performance across simulation and real-robot benchmarks spanning single-arm, bimanual, and dexterous manipulation. 
It demonstrates strong robustness to perturbations, achieving success rates of 94.8\% on LIBERO-Plus and 72.4\% on RoboTwin 2.0 C2R, alongside strong cross-embodiment transfer and real-world manipulation performance on Bench2Dex and RoboDojo-RealWorld, respectively. 
Pretraining-scale analysis shows larger performance gains from data scaling than a video-VAE-only baseline, indicating stronger scaling benefits and more effective data utilization. 
Moreover, Phoenix maintains high success rates when adapted with limited downstream demonstrations, demonstrating strong few-shot adaptation.
Our code, pretraining data, and models are available on the \href{https://openphoenix-team.github.io/Phoenix-pages/}{project page} to help the community reproduce our results.
\end{abstract}

\begin{figure}[t]
  \centering
  \includegraphics[width=0.95\linewidth]{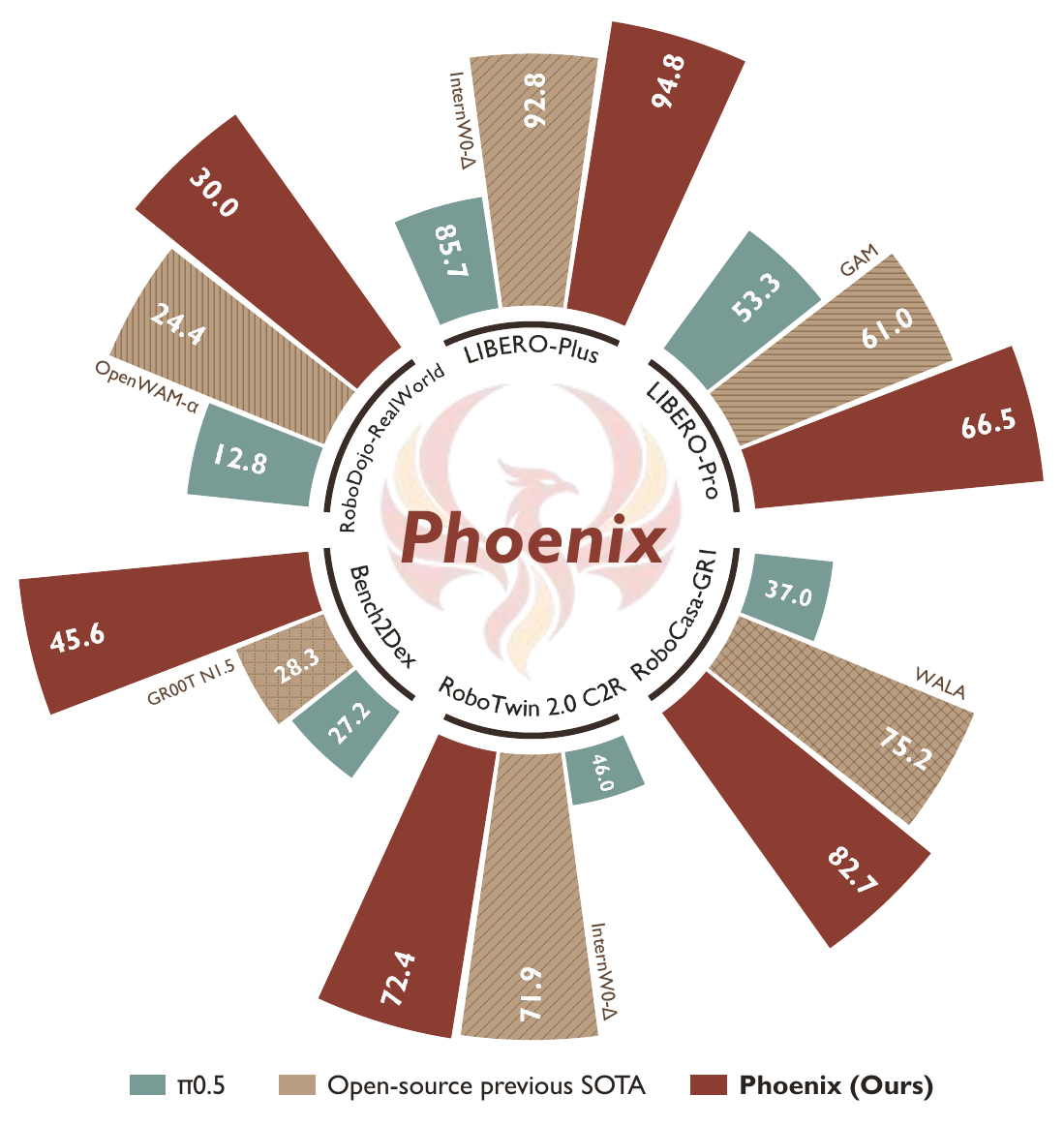}
  \caption{Phoenix generalizes across environments and embodiments.
  We report success rates (\%) on six simulation and real-world benchmarks after benchmark-specific adaptation.
  }
  \label{fig:teaser}
\end{figure}

\section{Introduction}
\label{sec:introduction}

World--action models (WAMs) jointly predict future observations and robot actions,
linking environment modeling with control~\cite{uva2025,dreamzero2026,openwam}.
They learn from large-scale visual experience to develop manipulation skills that
generalize across scenes, object configurations, and execution
conditions~\cite{vpp2025,genie2025,geact2026}.
This requires visual representations that support both environment prediction and policy learning.

In WAMs, visual representations determine how observations are encoded and what the model predicts. Manipulation-centered representations of \emph{scene structure} should capture spatial organization, visual appearance, and motion. Latents learned for video reconstruction preserve rich visual information, but reconstruction fidelity alone does not ensure that they capture manipulation-relevant structure and its relationship to actions~\cite{videoldm2023,wan,lingbotva2026}. Recent work on representation alignment and generation with visual foundation model features shows that visual understanding complements generative modeling~\cite{repa2025,rw_rae,dreamworld2026}. We therefore investigate how geometric and spatiotemporal semantic knowledge from these models~\cite{dino,omega,vjepa21} can be used to build scene representations for joint prediction of future states and actions.

We introduce \textbf{Phoenix}, a WAM that learns generative representations of scene structure. Phoenix Encoder builds on DINOv3's general visual semantics and learns per-frame scene features through dual-teacher distillation of geometric and spatiotemporal semantic knowledge. CS-VAE aggregates Phoenix Encoder features over causal temporal windows and compresses them into generative latents, preserving scene information and spatial layout. Phoenix jointly predicts these scene latents, complementary appearance latents, and robot actions, learning how actions relate to scene changes. To study pretraining data efficiency, we curate approximately \textbf{39,000 hours of egocentric and robot manipulation data} for representation learning and use the robot demonstration subset for WAM pretraining.

As shown in Figure~\ref{fig:teaser}, after benchmark-specific post-training, Phoenix achieves \textbf{state-of-the-art performance across six benchmarks}: LIBERO-Plus~\cite{plus}, LIBERO-Pro~\cite{pro}, RoboTwin 2.0 C2R~\cite{twin}, RoboCasa-GR1~\cite{casa}, Bench2Dex~\cite{dex}, and RoboDojo-RealWorld~\cite{dojo}. These benchmarks assess manipulation generalization under environmental perturbations and across robot embodiments, spanning simulation and real-world settings.

We further study how representations of scene structure affect manipulation performance and data efficiency across WAM pretraining and downstream adaptation.
Ablations show that learning these representations and compressing them into generative latents both improve manipulation performance,
accompanied by more faithful predictions of interaction dynamics.
Scaling WAM pretraining data yields larger gains in success rate for Phoenix than for a video-VAE-only baseline, indicating improved pretraining data efficiency. These benefits extend to downstream adaptation, where pretrained Phoenix maintains high success rates with limited task-specific demonstrations, showing effective knowledge transfer in \textbf{few-shot settings}.

Our main contributions are as follows:
\begin{itemize}
\item \textbf{Representations of scene structure for world--action modeling.} Phoenix Encoder builds on general visual semantics to learn features of scene structure through dual-teacher distillation of geometric and spatiotemporal semantic knowledge. CS-VAE causally aggregates and compresses them into generative latents, preserving spatial layout. Phoenix jointly predicts these latents, complementary appearance latents, and robot actions.
\item \textbf{Improved WAM pretraining data efficiency.} We compare visual representations across WAM pretraining scales using robot demonstrations and show that our representation yields larger performance gains from scaling data than a video-VAE-only baseline.
\item \textbf{State-of-the-art performance across six simulation and real-world benchmarks.} Phoenix achieves SOTA results in perturbation robustness, cross-embodiment adaptation, and dexterous and real-world manipulation. Further experiments demonstrate strong few-shot adaptation with limited task-specific demonstrations.
\end{itemize}

\section{Related Work}
\paragraph{World--Action Models and Visual Representations.}
World--action models connect future prediction with control through predictive video features~\cite{vpp2025} or joint video--action learning~\cite{uva2025,rw_uwm,dreamzero2026}. Video co-training can improve policies even without future generation at inference~\cite{fastwam}, highlighting the role of the learned visual representation.

Pretrained visual features enrich this prediction space with semantic knowledge. Such features have supported action-conditioned prediction and planning~\cite{rw_dinowm,assran2025vjepa2}, and can guide generative learning through alignment~\cite{repa2025} or serve as the latent space for generation~\cite{rw_rae}. Within WAMs, concurrent studies examine predictive feature learning and the integration of external representations~\cite{rw_rewam,rw_r2wam}. Recent approaches increasingly extend reconstruction-oriented visual representations with semantic knowledge, using pretrained semantic supervision~\cite{geact2026,rw_repwam,lingbotva2026} or predicting semantic features alongside appearance~\cite{dreamworld2026,rw_stwam}. A controlled comparison of reconstruction and semantic latent spaces~\cite{rw_reconstruction_semantics} further shows that reconstruction fidelity alone does not determine planning quality, and that semantic latents can better preserve action-relevant information. Phoenix learns representations of scene structure by enriching general visual semantics with geometric and spatiotemporal semantic knowledge through dual-teacher distillation. CS-VAE causally aggregates and compresses these Phoenix features into compact generative latents while preserving spatial layout. The Phoenix WAM jointly predicts these latents, complementary appearance latents, and robot actions.

\begin{figure*}[t]
 \centering
 \includegraphics[width=0.93\textwidth]{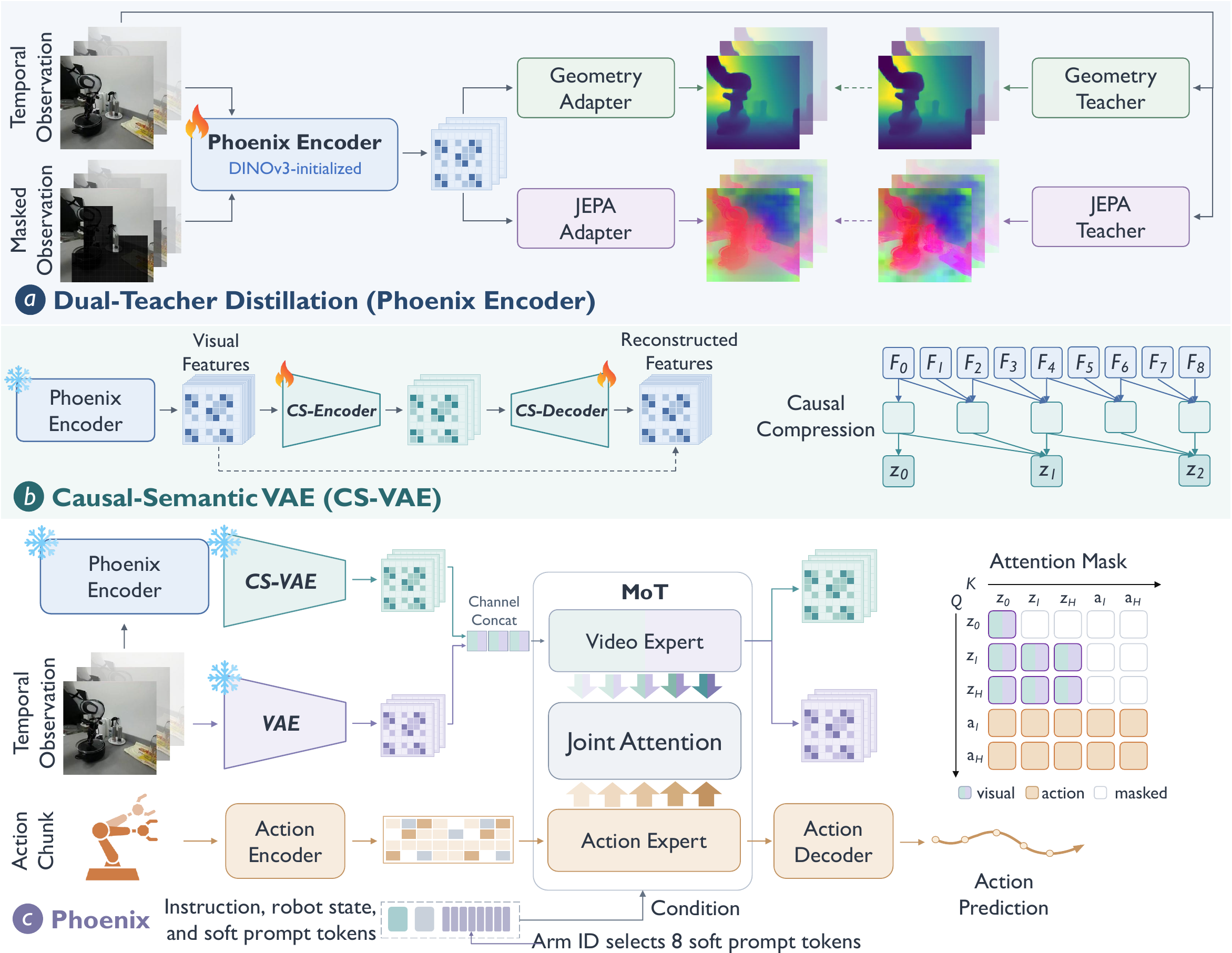}
 \caption{Phoenix architecture. (a) VGGT-$\Omega$ and V-JEPA~2.1-L distill geometry and spatiotemporal semantics into a DINOv3-based Phoenix Encoder. (b) CS-VAE reconstructs frozen Phoenix Encoder features while causally aggregating and compressing nine frames into three latent time steps, retaining the patch layout. (c) Action tokens attend to visual tokens; visual tokens cannot attend to actions. The experts jointly predict scene and appearance latents and action chunks. Soft prompt tokens condition the shared control interface.}
 \label{fig:method}
\end{figure*}

\paragraph{Embodied Pretraining from Heterogeneous Data.}
Large-scale pretraining pools data across tasks, scenes, and embodiments to support generalist robot policies~\cite{openvla,pi05}. For world--action learning, heterogeneous human and robot recordings provide complementary visual and action supervision. VPP2~\cite{rw_vpp2} develops a manipulation-video prior before visual-planner distillation and action learning. UWM~\cite{rw_uwm} and OpenWAM~\cite{openwam} study joint learning from heterogeneous data, while LDA-1B~\cite{supp_lda} uses recordings with different supervision to learn dynamics, visual forecasting, and policies. EgoWAM~\cite{rw_egowam} examines prediction targets for transferring from egocentric human videos to robot manipulation. Phoenix uses approximately 39,000 hours of egocentric and robot manipulation data for representation learning, followed by world--action pretraining on robot demonstrations. We compare representations across pretraining scales to assess how effectively additional data improves manipulation performance.

\section{Method}
Phoenix predicts action chunks and future visual latents from current visual observations, robot state, and a language instruction (Fig.~\ref{fig:method}). Phoenix Encoder learns per-frame scene features through dual-teacher distillation. CS-VAE, a causal-semantic variational autoencoder, aggregates and compresses these Phoenix features into generative latents. Together with complementary Wan VAE appearance latents, they support scene prediction and action learning. Training proceeds through representation pretraining on heterogeneous recordings, WAM pretraining on robot demonstrations, and task-specific post-training.

\subsection{Dual-teacher Encoder Distillation}
We initialize a ViT-L/16 student from the DINOv3 backbone in VGGT-$\Omega$~\cite{dino,omega}, building on its pretrained visual semantics. Phoenix Encoder learns per-frame scene representations by distilling geometric knowledge from VGGT-$\Omega$ and spatiotemporal semantic knowledge from V-JEPA~2.1-L. Two spatiotemporal adapters connect the per-frame backbone to clip-level supervision: a geometry adapter and a JEPA adapter.

For geometric distillation, we align intermediate per-frame and cross-view features with those of the frozen VGGT-$\Omega$ teacher:
\begin{equation}
\mathcal L_{\mathrm{align}}
=
\sum_{s\in\mathcal S}
\lambda_s\,\Delta(\widehat C_s,C_s),
\end{equation}
where $\widehat C_s$ and $C_s$ denote projected student features and teacher targets. The set $\mathcal S$ indexes supervised feature groups across teacher layers, $\Delta$ is the mean smooth-$L_1$ loss, and $\lambda_s$ weights each group. A confidence-weighted log-depth loss additionally matches teacher-generated depth, giving $\mathcal L_{\mathrm{geo}}=\mathcal L_{\mathrm{align}}+\mathcal L_{\mathrm{depth}}$ without ground-truth depth labels.

For spatiotemporal semantic distillation, the frozen V-JEPA~2.1-L teacher~\cite{vjepa21} observes the full clip, while the student receives spatial masks applied consistently across frames. A predictor maps the JEPA adapter outputs into the teacher feature space:
\begin{equation}
\begin{aligned}
\mathcal L_{\mathrm{tmp}}
={}&
\frac{1}{|\mathcal M|}
\sum_{i\in\mathcal M}\Delta(Q_i,Z_i)\\
&+
\frac{\lambda_{\mathrm{ctx}}}{|\mathcal V|}
\sum_{i\in\mathcal V}\alpha_i\,\Delta(Q_i,Z_i),
\end{aligned}
\end{equation}
where $\mathcal M$ and $\mathcal V$ denote masked and visible locations, and $Q_i$ and $Z_i$ are student predictions and teacher targets. Both terms use the same alignment loss. At visible locations, $\lambda_{\mathrm{ctx}}$ weights the loss, and the context weights $\alpha_i$ decrease with distance to the nearest masked location.

We alternate updates from the geometry and video teachers, following their respective pretraining objectives. Only the exponential moving average (EMA) backbone is retained for policy training and deployment. Full objectives and training details appear in the \textit{supplementary material}.

\subsection{Causal Aggregation and Compression}
Phoenix Encoder produces dense, high-dimensional per-frame scene features. CS-VAE compresses their channel and temporal dimensions while aggregating features over causal temporal windows and retaining the spatial patch grid. Feature reconstruction preserves scene information, while KL regularization encourages the latent posterior to match a standard Gaussian prior.

CS-VAE applies a spatial Transformer per frame, followed by two causal temporal compression stages and a projection that parameterizes a Gaussian latent posterior. It compresses nine feature frames $F_0,\dots,F_8$ into three latent frames $z_0,z_1,z_2$, preserving the spatial patch grid.

The first latent $z_0$ depends only on $F_0$. Each subsequent latent aggregates a causal window of consecutive Phoenix feature frames ending at its corresponding time step, extending temporal context beyond individual frames. This preserves the current observation as conditioning while organizing the remaining sequence into temporally aggregated prediction targets.

During CS-VAE training, we sample latent codes $z$ from the approximate Gaussian posterior $q(z\mid F)$ parameterized by the CS-VAE encoder, where $F$ denotes frozen Phoenix Encoder features after fixed channel-wise standardization. The decoder $D$ reconstructs the feature sequence from these codes, yielding $\hat F=D(z)$. The CS-VAE objective is
\begin{equation}
\begin{split}
 \mathcal L_{\text{CS-VAE}} ={}&
 \mathcal L_{\mathrm{MSE}}(\hat F_t,F_t)
 +\mathcal L_{\mathrm{cos}}(\hat F_t,F_t)\\
 &+\beta D_{\mathrm{KL}}\!\left(q(z\mid F)\,\|\,\mathcal N(0,I)\right).
\end{split}
\end{equation}
Here, $F_t$ and $\hat F_t$ denote the target and reconstructed features at frame $t$. $\mathcal L_{\mathrm{MSE}}$ and $\mathcal L_{\mathrm{cos}}$ are the mean squared error and cosine reconstruction losses. The KL divergence $D_{\mathrm{KL}}$, weighted by $\beta$, regularizes the posterior toward the standard Gaussian prior $\mathcal N(0,I)$, with identity covariance $I$. The resulting latents complement video-VAE appearance latents for world--action modeling. Training details are provided in the \textit{supplementary material}.

\subsection{Joint Visual and Action Modeling}
We concatenate CS-VAE scene latents with Wan VAE appearance latents along channels. A video diffusion Transformer models the concatenated visual tokens, while an action diffusion Transformer models action chunks. Masked joint attention couples the two streams at each layer with an asymmetric mask. Current-frame visual tokens attend only to one another; future visual tokens attend to all visual tokens but not actions; and action tokens attend to all visual and action tokens. The remaining computation is modality-specific. The action stream also cross-attends to T5 instruction embeddings, robot states, and soft prompt tokens indexed by arm.

We train the coupled experts with flow matching~\cite{flow}: the current visual latent stays clean, while the future latents and the action chunk are perturbed with Gaussian noise, so each expert regresses its own flow field. The objective is
\begin{equation}
 \mathcal L_{\mathrm{WAM}} =
 0.5\,\mathcal L_{\rm appearance}
 +0.5\,\mathcal L_{\rm feature}
 +\mathcal L_{\rm action}.
\end{equation}
$\mathcal L_{\rm appearance}$ supervises the appearance latents, $\mathcal L_{\rm feature}$ the future CS-VAE latents, and $\mathcal L_{\rm action}$ the action chunk, scored only over the control dimensions the embodiment provides.

At inference, we fix current visual latents and jointly denoise future latents and action chunks from Gaussian noise.

\newsavebox{\benchmarkmeasure}
\begin{table*}[t]
\centering\small
\caption{LIBERO-Plus and RoboTwin 2.0 Clean2Random success rates.}\label{tab:plus}\label{tab:twin}
\setlength{\tabcolsep}{0pt}
\sbox{\benchmarkmeasure}{\begin{tabular}{lccccccccc}\toprule
 & \multicolumn{8}{c}{LIBERO-Plus} & RoboTwin 2.0\\
\cmidrule(lr){2-9}\cmidrule(lr){10-10}
Method & Camera & Robot & Language & Light & Background & Noise & Layout & Total $\uparrow$ & Clean2Random\\ \midrule
\rowcolor{benchmarkgroup}\multicolumn{10}{c}{\textit{VLA}}\\
OpenVLA~\cite{openvla} & 0.8 & 3.5 & 23.0 & 8.1 & 34.8 & 15.2 & 28.5 & 15.6 & --\\
$\pi_0$~\cite{pi0} & 13.8 & 6.0 & 58.8 & 85.0 & 81.4 & 79.0 & 68.9 & 53.6 & 16.3\\
OpenVLA-OFT~\cite{oft} & 56.4 & 31.9 & 79.5 & 88.7 & 93.3 & 75.8 & 74.2 & 69.6 & --\\
StarVLA~\cite{supp_starvla} & 47.0 & 60.1 & 87.0 & 96.3 & 95.3 & 73.1 & 79.2 & 75.0 & 10.6\\
ABot-M0~\cite{supp_abot0} & 60.4 & 67.9 & 86.4 & 96.2 & 91.6 & 86.4 & 82.6 & 80.5 & 36.0\\
$\pi_{0.5}$~\cite{pi05} & 75.8 & 79.4 & 83.3 & 95.5 & 95.0 & 89.6 & 87.0 & 85.7 & 46.0\\
Qwen-RobotManip-Context~\cite{qwen} & \underline{89.9} & 83.9 & 86.5 & \underline{98.6} & \textbf{99.9} & \underline{97.9} & 87.5 & 91.4 & 69.4\\
\rowcolor{benchmarkgroup}\multicolumn{10}{c}{\textit{WAM}}\\
Fast-WAM~\cite{fastwam} & 16.4 & 44.5 & 68.9 & 78.2 & 53.7 & 37.7 & 60.7 & 51.5 & 1.9\\
OpenWAM-$\alpha$~\cite{openwam} & 33.8 & 76.1 & 88.0 & 97.0 & 87.1 & 39.8 & 77.5 & 69.2 & 48.7\\
Cosmos-Policy~\cite{cosmos} & 75.8 & 63.3 & 81.7 & 96.5 & 88.9 & 92.7 & 82.2 & 82.2 & --\\
ImageWAM~\cite{imagewam} & 80.8 & 50.3 & 91.4 & 98.1 & 85.5 & 93.8 & 80.5 & 83.1 & --\\
InternW0-$\Delta$~\cite{intern} & \textbf{90.6} & \underline{91.1} & \underline{92.9} & 95.8 & \underline{99.4} & 94.0 & \underline{88.4} & \underline{92.8} & \underline{71.9}\\
\rowcolor{phoenixrow}\textbf{Phoenix (Ours)} & \underline{89.9} & \textbf{94.0} & \textbf{94.7} & \textbf{99.9} & 98.6 & \textbf{98.7} & \textbf{90.0} & \textbf{94.8} & \textbf{72.4}\\
\bottomrule\end{tabular}}
\setlength{\tabcolsep}{\dimexpr(\linewidth-\wd\benchmarkmeasure)/20\relax}
\typeout{BENCHMARK merged width=\the\linewidth; natural=\the\wd\benchmarkmeasure; tabcolsep=\the\tabcolsep}
\scalebox{0.94}{%
\begin{tabular}{lccccccccc}\toprule
 & \multicolumn{8}{c}{LIBERO-Plus} & RoboTwin 2.0\\
\cmidrule(lr){2-9}\cmidrule(lr){10-10}
Method & Camera & Robot & Language & Light & Background & Noise & Layout & Total $\uparrow$ & Clean2Random\\ \midrule
\rowcolor{benchmarkgroup}\multicolumn{10}{c}{\textit{VLA}}\\
OpenVLA~\cite{openvla} & 0.8 & 3.5 & 23.0 & 8.1 & 34.8 & 15.2 & 28.5 & 15.6 & --\\
$\pi_0$~\cite{pi0} & 13.8 & 6.0 & 58.8 & 85.0 & 81.4 & 79.0 & 68.9 & 53.6 & 16.3\\
OpenVLA-OFT~\cite{oft} & 56.4 & 31.9 & 79.5 & 88.7 & 93.3 & 75.8 & 74.2 & 69.6 & --\\
StarVLA~\cite{supp_starvla} & 47.0 & 60.1 & 87.0 & 96.3 & 95.3 & 73.1 & 79.2 & 75.0 & 10.6\\
ABot-M0~\cite{supp_abot0} & 60.4 & 67.9 & 86.4 & 96.2 & 91.6 & 86.4 & 82.6 & 80.5 & 36.0\\
$\pi_{0.5}$~\cite{pi05} & 75.8 & 79.4 & 83.3 & 95.5 & 95.0 & 89.6 & 87.0 & 85.7 & 46.0\\
Qwen-RobotManip-Context~\cite{qwen} & \underline{89.9} & 83.9 & 86.5 & \underline{98.6} & \textbf{99.9} & \underline{97.9} & 87.5 & 91.4 & 69.4\\
\rowcolor{benchmarkgroup}\multicolumn{10}{c}{\textit{WAM}}\\
Fast-WAM~\cite{fastwam} & 16.4 & 44.5 & 68.9 & 78.2 & 53.7 & 37.7 & 60.7 & 51.5 & 1.9\\
OpenWAM-$\alpha$~\cite{openwam} & 33.8 & 76.1 & 88.0 & 97.0 & 87.1 & 39.8 & 77.5 & 69.2 & 48.7\\
Cosmos-Policy~\cite{cosmos} & 75.8 & 63.3 & 81.7 & 96.5 & 88.9 & 92.7 & 82.2 & 82.2 & --\\
ImageWAM~\cite{imagewam} & 80.8 & 50.3 & 91.4 & 98.1 & 85.5 & 93.8 & 80.5 & 83.1 & --\\
InternW0-$\Delta$~\cite{intern} & \textbf{90.6} & \underline{91.1} & \underline{92.9} & 95.8 & \underline{99.4} & 94.0 & \underline{88.4} & \underline{92.8} & \underline{71.9}\\
\rowcolor{phoenixrow}\textbf{Phoenix (Ours)} & \underline{89.9} & \textbf{94.0} & \textbf{94.7} & \textbf{99.9} & 98.6 & \textbf{98.7} & \textbf{90.0} & \textbf{94.8} & \textbf{72.4}\\
\bottomrule\end{tabular}
}
\end{table*}

\subsection{Pretraining and Post-training}

\paragraph{Pretraining data.}
We filter 54,413 hours of data and retain 39,306 hours for pretraining.
Robot demonstrations, egocentric videos, and Universal Manipulation
Interface (UMI) data account for 31.09\%, 63.58\%, and 5.33\% of the
retained duration, respectively.

\paragraph{Representation pretraining.}
Stages (a) and (b) in Fig.~\ref{fig:method} use all filtered data.
In stage (a), we distill both frozen teachers into Phoenix Encoder to learn scene features.
In stage (b), CS-VAE learns causal aggregation and generative compression of frozen Phoenix Encoder features.

\paragraph{World--action model pretraining.}
Stage (c) in Fig.~\ref{fig:method} uses robot demonstrations to jointly
predict future visual latents and action chunks.
For cross-embodiment training, we map robot states and actions
into a shared 80-dimensional interface.
Each arm ID selects eight learnable soft prompt tokens, which condition
the action expert together with language and robot state.
We freeze Phoenix Encoder, CS-VAE, and the video VAE, and update
the video and action experts, state projection, and soft prompt tokens.

\paragraph{Post-training.}
We adapt the pretrained model using target-task robot demonstrations,
retaining the same trainable and frozen modules as in world--action
pretraining.

\section{Experiments}
\subsection{Implementation Details}
\paragraph{Network architecture.}
CS-VAE causally aggregates and compresses nine frames of Phoenix Encoder features into 48-channel scene latents with three temporal positions. These are concatenated with the VAE's 48-channel appearance latents to form a 96-channel visual representation. The video expert is initialized from Wan2.2-TI2V-5B, and the action expert is initialized randomly. Each predicted action chunk spans 32 consecutive control steps.

\paragraph{Data settings.}
The per-view input resolution is $256\times256$ throughout training. Encoder distillation and CS-VAE training use filtered robot demonstrations, egocentric videos, and UMI recordings; world--action pretraining uses only robot demonstrations. For robot demonstrations from different sources, we standardize end-effector pose representations through workspace alignment, end-effector alignment, and consistent position units. Data-scaling experiments sample complete trajectories. Pretraining groups trajectories by task combinations to form the 50\% subset $\mathcal{D}_{50}$, then subsamples it according to full-dataset source proportions to form $\mathcal{D}_{25}$ (25\% of the full dataset), with \mbox{$\mathcal{D}_{25}\subset\mathcal{D}_{50}\subset\mathcal{D}_{100}$}. Post-training subsets are sampled progressively within each task, satisfying \mbox{$\mathcal{D}_{25}\subset\mathcal{D}_{50}\subset\mathcal{D}_{75}\subset\mathcal{D}_{100}$} while retaining all tasks.

\paragraph{Training and inference.}
Encoder distillation uses cosine decay with peak learning rates of $10^{-5}$ for the backbone and $10^{-4}$ for adapters/predictor; subsequent stages use the 110k-step EMA encoder checkpoint. CS-VAE trains for 150k steps with a peak learning rate of $1.5\times10^{-4}$. World--action pretraining freezes Phoenix Encoder, CS-VAE, and Wan VAE, and uses AdamW with a global batch size of 4,480 and a peak learning rate of $10^{-4}$ for one epoch at each data scale. Post-training runs for 25k steps by default, and policy inference uses 10 denoising steps.



\subsection{Generalizable Manipulation across Perturbations and Embodiments}
After WAM pretraining and benchmark-specific post-training, we evaluate Phoenix's generalization across environmental perturbations and robot embodiments.

\paragraph{Generalization under perturbations.}
LIBERO-Plus evaluates seven perturbation categories spanning visual appearance, robot initialization, language, and object layout. Phoenix achieves 94.8\% aggregate success, exceeding InternW0-$\Delta$ by 2.0 percentage points (Table~\ref{tab:plus}), and leads in five of the seven categories. Extending the evaluation to bimanual manipulation, RoboTwin 2.0 Clean2Random~\cite{twin} tests transfer from clean training scenes to randomized environments. Phoenix achieves 72.4\% success, surpassing InternW0-$\Delta$ by 0.5 percentage points and $\pi_{0.5}$ by 26.4 points (Table~\ref{tab:twin}). These results support learning scene representations for world--action modeling to maintain task performance across changing manipulation conditions.

\begin{table}[t]
\centering\small\setlength{\tabcolsep}{2.5pt}
\caption{Bench2Dex success rate.}\label{tab:dex}
\setlength{\tabcolsep}{0pt}
\sbox{\benchmarkmeasure}{\begin{tabular}{lrrrrr}\toprule
Method & None & Equi. & Inv. & Full & SR\\ \midrule
\rowcolor{benchmarkgroup}\multicolumn{6}{c}{\textit{Visuomotor Policy}}\\
ACT~\cite{act} & 29.5 & 15.4 & 14.0 & 13.0 & 18.0\\
Diffusion Policy~\cite{dp} & 12.9 & 6.3 & 5.2 & 3.8 & 7.1\\
\rowcolor{benchmarkgroup}\multicolumn{6}{c}{\textit{VLA}}\\
$\pi_{0.5}$~\cite{pi05} & 35.5 & \underline{21.2} & \underline{26.7} & \underline{25.5} & 27.2\\
GR00T N1.5~\cite{gr00tn1_2025} & \underline{48.5} & 18.8 & 26.0 & 19.8 & \underline{28.3}\\
\rowcolor{benchmarkgroup}\multicolumn{6}{c}{\textit{WAM}}\\
\rowcolor{phoenixrow}\textbf{Phoenix (Ours)} & \textbf{54.2} & \textbf{38.5} & \textbf{44.1} & \textbf{45.6} & \textbf{45.6}\\
\bottomrule\end{tabular}}
\setlength{\tabcolsep}{\dimexpr(\linewidth-\wd\benchmarkmeasure)/12\relax}
\typeout{BENCHMARK dex width=\the\linewidth; tabcolsep=\the\tabcolsep}
\scalebox{0.98}{%
\begin{tabular}{lrrrrr}\toprule
Method & None & Equi. & Inv. & Full & SR\\ \midrule
\rowcolor{benchmarkgroup}\multicolumn{6}{c}{\textit{Visuomotor Policy}}\\
ACT~\cite{act} & 29.5 & 15.4 & 14.0 & 13.0 & 18.0\\
Diffusion Policy~\cite{dp} & 12.9 & 6.3 & 5.2 & 3.8 & 7.1\\
\rowcolor{benchmarkgroup}\multicolumn{6}{c}{\textit{VLA}}\\
$\pi_{0.5}$~\cite{pi05} & 35.5 & \underline{21.2} & \underline{26.7} & \underline{25.5} & 27.2\\
GR00T N1.5~\cite{gr00tn1_2025} & \underline{48.5} & 18.8 & 26.0 & 19.8 & \underline{28.3}\\
\rowcolor{benchmarkgroup}\multicolumn{6}{c}{\textit{WAM}}\\
\rowcolor{phoenixrow}\textbf{Phoenix (Ours)} & \textbf{54.2} & \textbf{38.5} & \textbf{44.1} & \textbf{45.6} & \textbf{45.6}\\
\bottomrule\end{tabular}%
}
\end{table}

\begin{table}[t]
\centering\small\setlength{\tabcolsep}{3pt}
\caption{RoboDojo-RealWorld success rate.}\label{tab:real}
\setlength{\tabcolsep}{0pt}
\sbox{\benchmarkmeasure}{\begin{tabular}{lrrrr}\toprule
Method & ARX X5 & Piper & Piper X & Avg.\\ \midrule
\rowcolor{benchmarkgroup}\multicolumn{5}{c}{\textit{VLA}}\\
StarVLA~\cite{supp_starvla} & 0.0 & 5.0 & 0.0 & 1.7\\
$\pi_0$~\cite{pi0} & 0.0 & 5.0 & 0.0 & 1.7\\
X-VLA~\cite{xvla} & 1.7 & 8.3 & 0.0 & 3.3\\
$\pi_{0.5}$~\cite{pi05} & 13.3 & 21.7 & 3.3 & 12.8\\
\rowcolor{benchmarkgroup}\multicolumn{5}{c}{\textit{WAM}}\\
OpenWAM-$\alpha$~\cite{openwam} & \underline{23.3} & \textbf{36.7} & \underline{13.3} & \underline{24.4}\\
\rowcolor{phoenixrow}\textbf{Phoenix (Ours)} & \textbf{33.3} & \underline{31.7} & \textbf{25.0} & \textbf{30.0}\\
\bottomrule\end{tabular}}
\setlength{\tabcolsep}{\dimexpr(\linewidth-\wd\benchmarkmeasure)/10\relax}
\typeout{BENCHMARK real width=\the\linewidth; tabcolsep=\the\tabcolsep}
\scalebox{0.97}{%
\begin{tabular}{lrrrr}\toprule
Method & ARX X5 & Piper & Piper X & Avg.\\ \midrule
\rowcolor{benchmarkgroup}\multicolumn{5}{c}{\textit{VLA}}\\
StarVLA~\cite{supp_starvla} & 0.0 & 5.0 & 0.0 & 1.7\\
$\pi_0$~\cite{pi0} & 0.0 & 5.0 & 0.0 & 1.7\\
X-VLA~\cite{xvla} & 1.7 & 8.3 & 0.0 & 3.3\\
$\pi_{0.5}$~\cite{pi05} & 13.3 & 21.7 & 3.3 & 12.8\\
\rowcolor{benchmarkgroup}\multicolumn{5}{c}{\textit{WAM}}\\
OpenWAM-$\alpha$~\cite{openwam} & \underline{23.3} & \textbf{36.7} & \underline{13.3} & \underline{24.4}\\
\rowcolor{phoenixrow}\textbf{Phoenix (Ours)} & \textbf{33.3} & \underline{31.7} & \textbf{25.0} & \textbf{30.0}\\
\bottomrule\end{tabular}%
}
\end{table}

\paragraph{Generalization across embodiments and configurations.}
Bench2Dex~\cite{dex} evaluates 26 task--embodiment settings spanning 12 arm--hand combinations. Phoenix achieves 45.6\% aggregate success, outperforming GR00T N1.5 by 17.3 percentage points (Table~\ref{tab:dex}), and ranks first across unperturbed, equivariant, invariant, and combined conditions. We further evaluate adaptation on physical robot platforms using RoboDojo-RealWorld~\cite{dojo}, which covers ARX X5, Piper, and Piper X. Phoenix achieves 33.3\%, 31.7\%, and 25.0\% success, respectively, leading on ARX~X5 and Piper~X. Its 30.0\% mean exceeds OpenWAM-$\alpha$ by 5.6 percentage points and $\pi_{0.5}$ by 17.2 points (Table~\ref{tab:real}). These results demonstrate the adaptability of the pretrained Phoenix framework across diverse robot configurations and manipulation tasks in both simulation and the real world.

Phoenix also achieves state-of-the-art performance on LIBERO-Pro~\cite{pro} and RoboCasa-GR1~\cite{casa}, with success rates of 66.5\% and 82.7\%, respectively. Full comparisons are provided in the \textit{supplementary material}.

\subsection{Ablation Study and In-depth Analysis}
\paragraph{Q1: Does our representation design support generalizable manipulation?}
\label{sec:spatial-probing}
Phoenix Encoder distills geometric and spatiotemporal semantic knowledge into pretrained visual features to learn scene representations. CS-VAE causally aggregates and compresses these features into generative latents. We evaluate these representations through frozen-feature probing and downstream ablations.

We follow the Spatial State benchmark~\cite{yin2026probingspatialstaterepresentations,huang2025vidfm3d} and probe frozen encoder features on the ScanNet++~\cite{yeshwanthliu2023scannetpp} subset of InsScene-15K~\cite{li2025iggt}, using identical task-specific probe architectures and training schedules across encoders. \textit{Spatial readouts} assess metric depth and per-pixel instance identity using prediction losses, and cross-view consistency using balanced correspondence accuracy. \textit{Predictive readouts} assess direct future-feature prediction from relative camera motion and path integration by chaining relative motions, using Global R@1 within each encoder's feature space. Scores are averaged over observation-prefix lengths.

\begin{table}[t]
\centering\footnotesize\renewcommand{\arraystretch}{1.06}
\caption{Spatial and predictive probing results.}
\label{tab:spatial-probing}
\newcommand{\spatialprobingrows}{%
\toprule
 & \multicolumn{3}{c}{Spatial readout} & \multicolumn{2}{c}{Predictive readout} & Rank \\
\cmidrule(lr){2-4}\cmidrule(lr){5-6}\cmidrule(lr){7-7}
Model & Depth $\downarrow$ & Instance $\downarrow$ & View $\uparrow$ & Direct $\uparrow$ & Path $\uparrow$ & Avg. $\downarrow$ \\
\midrule
DINOv2-L~\cite{oquab2024dinov2learningrobustvisual} & 0.1422 & \textbf{1.3828} & 95.57 & \underline{23.73} & 9.00 & \underline{2.60} \\
Fast3R~\cite{Yang_2025_Fast3R} & \textbf{0.1386} & 1.6072 & \textbf{98.08} & 1.33 & 1.90 & 3.40 \\
VideoMAE-L~\cite{tong2022videomaemaskedautoencodersdataefficient} & 0.2968 & 4.1571 & 91.18 & 5.31 & 6.85 & 5.60 \\
V-JEPA~2-L~\cite{assran2025vjepa2} & 0.2708 & 2.9211 & 92.56 & 6.25 & \textbf{15.88} & 3.80 \\
Qwen2.5-VL~\cite{bai2025qwen25vltechnicalreport} & 0.2244 & 2.9476 & 93.85 & 17.12 & 9.03 & 3.80 \\
\midrule
\rowcolor{phoenixrow}\textbf{Phoenix Encoder} & \underline{0.1411} & \underline{1.5738} & \underline{96.18} & \textbf{34.42} & \underline{15.01} & \textbf{1.80} \\
\bottomrule
}
\setlength{\tabcolsep}{0pt}
\sbox{\benchmarkmeasure}{\begin{tabular}{lcccccc}\spatialprobingrows\end{tabular}}
\setlength{\tabcolsep}{\dimexpr(\linewidth-\wd\benchmarkmeasure)/14\relax}
\typeout{PROBING TABLE width=\the\linewidth; natural=\the\wd\benchmarkmeasure; tabcolsep=\the\tabcolsep}
\begin{tabular}{lcccccc}\spatialprobingrows\end{tabular}
\end{table}
\begin{figure}[t]
\centering
\includegraphics[width=0.95\columnwidth]{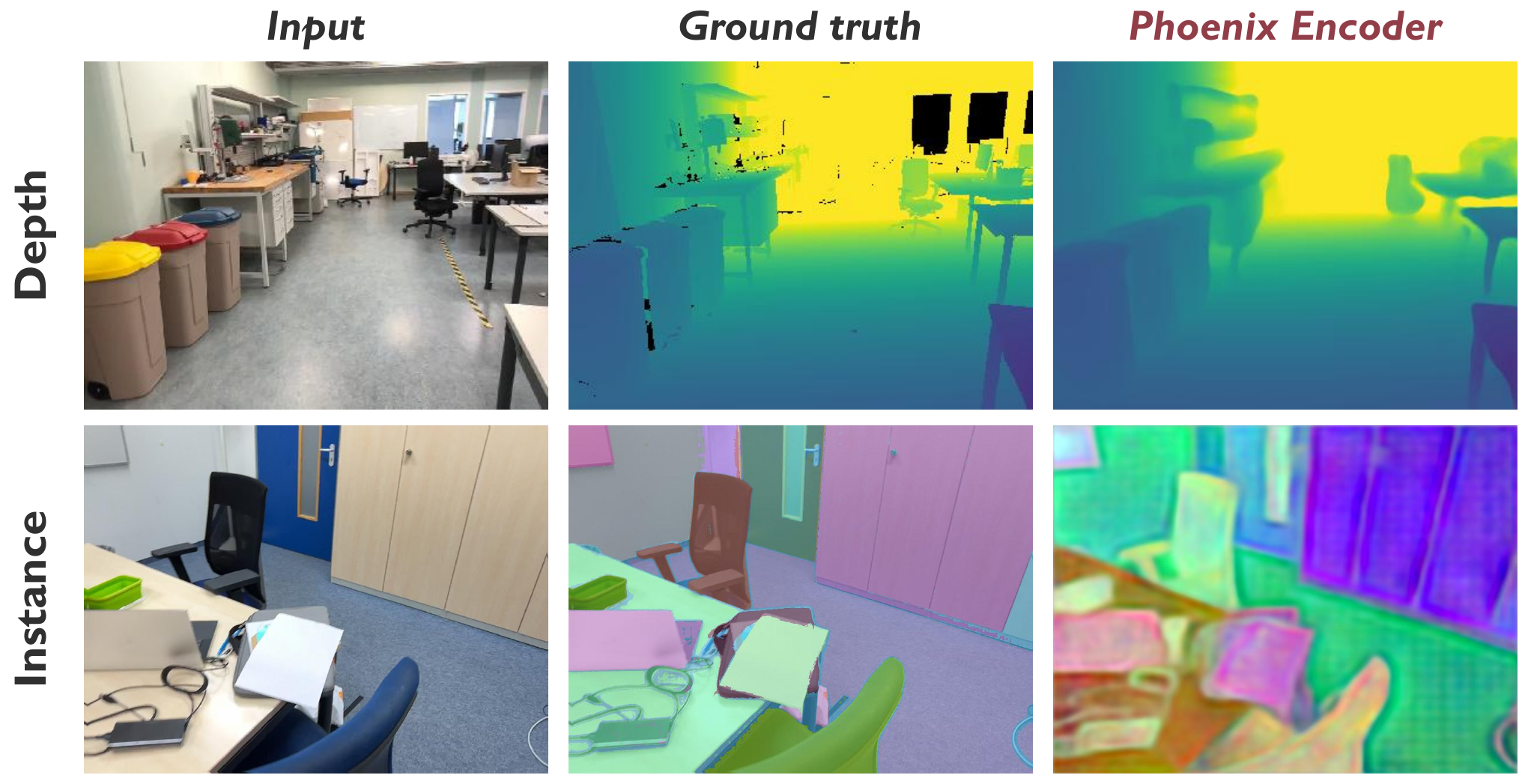}
\caption{Depth (top; 0--6\,m) and instance probing (bottom).
}
\label{fig:probing-qualitative}
\end{figure}
\begin{table*}[t]
\centering\small
\caption{Visual representation and feature compression ablations. All models are post-trained without WAM pretraining.}
\label{tab:representation-ablation}
\label{tab:compression-ablation}
\renewcommand{\arraystretch}{1.15}
\def\phoenixablationrows{%
\toprule
 & \multicolumn{2}{c}{Representation} & \multicolumn{8}{c}{LIBERO-Plus} \\
\cmidrule(lr){2-3}\cmidrule(lr){4-11}
Appearance & Encoder & Compression & Camera & Robot & Language & Light & Background & Noise & Layout & Total $\uparrow$ \\
\midrule
VAE & -- & -- & 75.5 & 69.0 & 81.9 & 84.7 & 55.0 & 88.3 & 79.0 & 76.9 \\
\midrule
VAE & VGGT-$\Omega$ & PCA48 & \textbf{81.9} & 60.4 & 64.5 & 94.2 & 85.0 & 90.0 & 77.9 & 78.3 \\
VAE & V-JEPA~2.1-L & PCA48 & 77.1 & \underline{72.3} & 76.9 & \textbf{98.7} & \underline{86.9} & 92.8 & 83.0 & 83.2 \\
VAE & DINOv3 & PCA48 & 61.7 & 66.2 & 68.3 & 88.2 & 79.2 & 86.1 & 76.6 & 74.4 \\
\midrule
VAE & Phoenix Encoder & PCA48 & 79.0 & 70.2 & \underline{82.8} & \underline{98.5} & 86.4 & \underline{94.4} & \underline{83.3} & \underline{84.4} \\
\rowcolor{phoenixrow}
VAE & Phoenix Encoder & CS-VAE & \underline{79.4} & \textbf{90.2} & \textbf{83.7} & 97.7 & \textbf{87.7} & \textbf{94.8} & \textbf{85.9} & \textbf{88.2} \\
\bottomrule
}
\setlength{\tabcolsep}{0pt}
\setbox0=\hbox{\begin{tabular}{llccccccccc}\phoenixablationrows\end{tabular}}
\dimen0=\dimexpr\textwidth-\wd0\relax
\divide\dimen0 by 22
\setlength{\tabcolsep}{\dimen0}
\scalebox{0.9}{%
\begin{tabular}{llccccccccc}
\phoenixablationrows
\end{tabular}%
}
\end{table*}

Phoenix Encoder achieves the best average rank (1.80), ranking first in direct
prediction (34.42\%) and second on the other four readouts
(Table~\ref{tab:spatial-probing}). The frozen representation balances spatial understanding
and motion-conditioned prediction. Figure~\ref{fig:probing-qualitative} shows depth and instance readouts.

We next evaluate visual representation design on LIBERO-Plus, retaining the appearance-VAE branch and fixing feature compression to PCA48. All models are post-trained without WAM pretraining (Table~\ref{tab:representation-ablation}). Phoenix Encoder achieves 84.4\% success, compared with 74.4\% for DINOv3 and 83.2\% for V-JEPA~2.1-L, the strongest alternative. This supports the benefit of learned scene features with compression held fixed.

With Phoenix Encoder fixed, replacing PCA48 with CS-VAE raises success from 84.4\% to 88.2\%. It improves six of seven perturbation categories, with the largest gain in robot initialization, from 70.2\% to 90.2\%. The full configuration leads in five categories and exceeds the VAE-only baseline by 11.3 points overall.
These results support complementary roles for scene feature learning and generative compression in manipulation generalization.

\begin{figure}[t]
\centering
\includegraphics[width=0.95\linewidth]{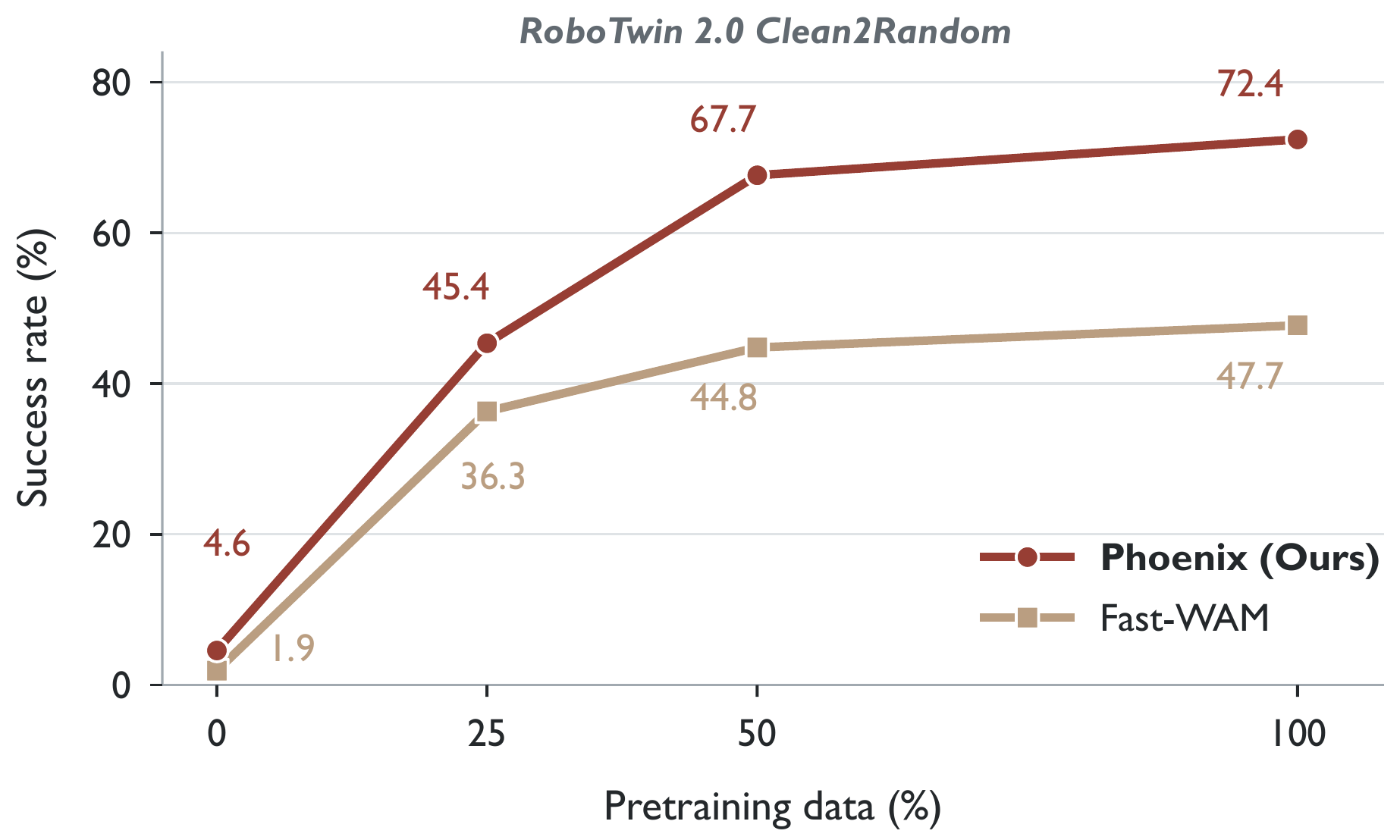}
\caption{Pretraining data scaling. Phoenix gains more from scaling pretraining data than the VAE-only baseline (Fast-WAM).}
\label{fig:pretraining-data-scaling}
\end{figure}
\vspace{-2.0mm}

\begin{figure}[t]
\centering
\includegraphics[width=0.95\linewidth]{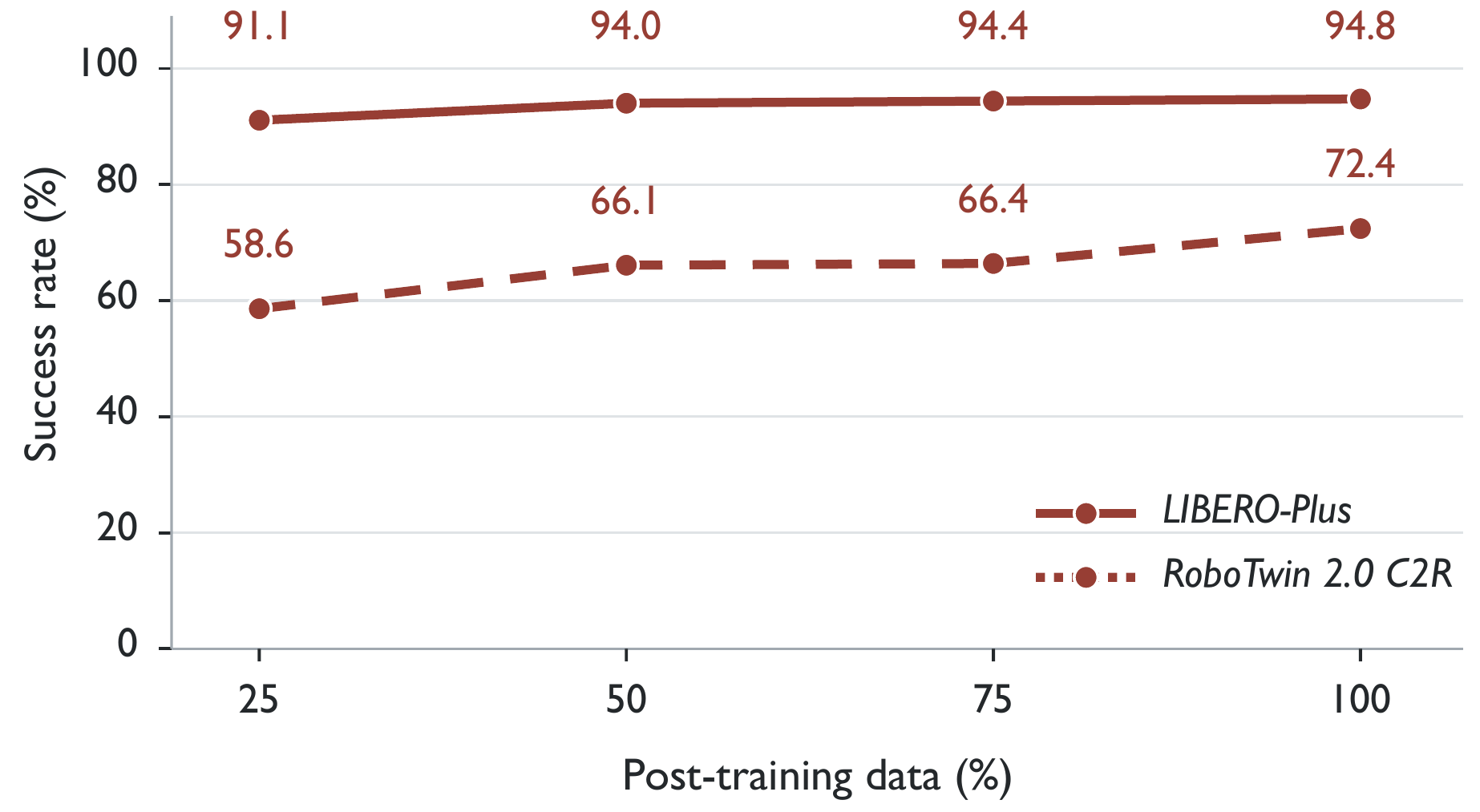}
\caption{Post-training data efficiency of pretrained Phoenix.}
\label{fig:posttraining-data-efficiency}
\end{figure}
\vspace{-2mm}
\begin{figure*}[!t]
\centering
\includegraphics[width=0.98\textwidth]{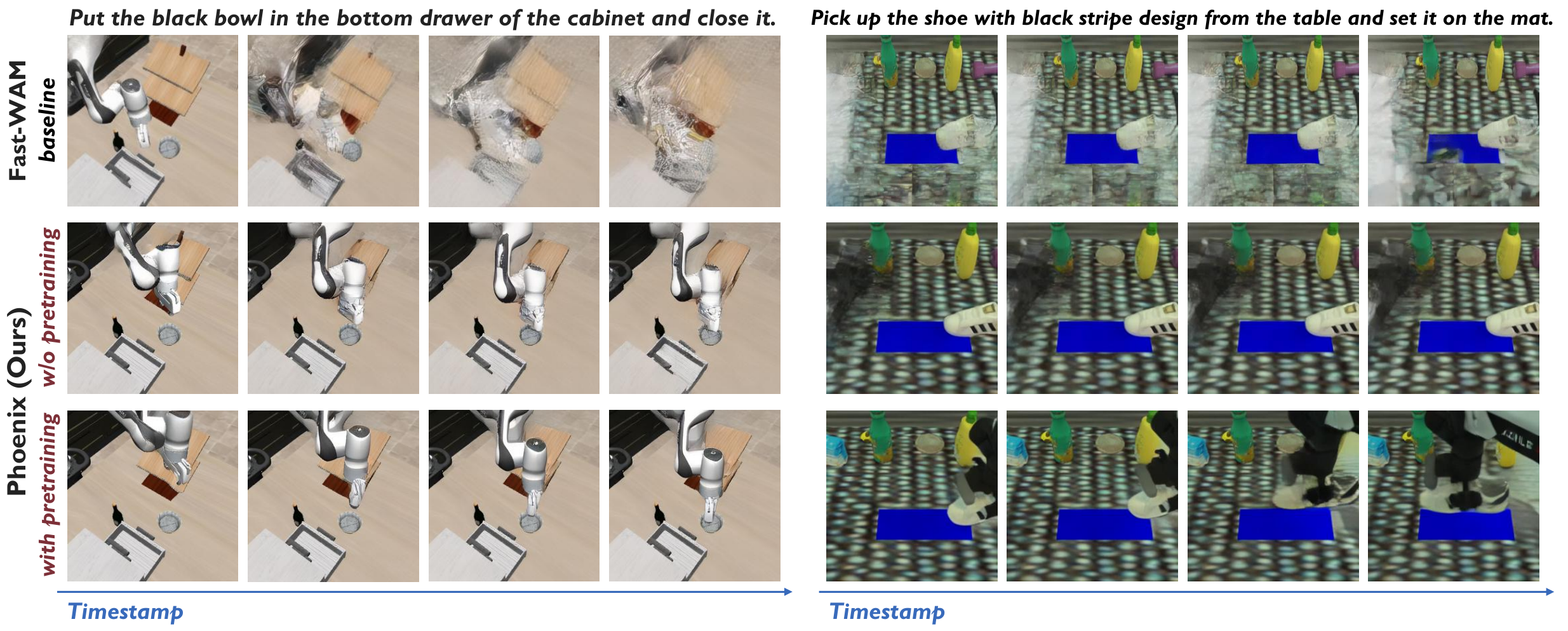}
\caption{Future-video predictions on LIBERO-Plus (left) and RoboTwin 2.0 C2R (right).
Phoenix preserves clearer object structure and more coherent interaction dynamics.
}
\label{fig:future-prediction}
\end{figure*}

\paragraph{Q2: Does our representation improve pretraining scaling and data efficiency?}
We examine how representation design affects the gains from scaling WAM pretraining data, with the learned scene representation held fixed. Figure~\ref{fig:pretraining-data-scaling} compares Phoenix with the VAE-only Fast-WAM baseline pretrained under our data-scaling protocol on RoboTwin 2.0 C2R. As the pretraining data fraction increases from 25\% to 100\%, Phoenix's success rate improves by 27.0 points, compared with 11.4 points for the baseline. Phoenix therefore gains more from additional pretraining data, widening its advantage over the baseline as the dataset grows.

Phoenix also achieves stronger performance with less pretraining data. Using 50\% of the WAM pretraining data, it reaches 67.7\% success, exceeding the baseline's 47.7\% with the full dataset. These results indicate that Phoenix's representation design helps learn transferable policies from manipulation experience, yielding larger gains from scaling pretraining data and higher data efficiency.

\paragraph{Q3: Can pretrained Phoenix adapt effectively with fewer demonstrations?}
We examine how well the pretrained model adapts with limited task-specific demonstrations.
Starting from the same checkpoint pretrained on the full pretraining dataset, we post-train separately on each benchmark using 25\%, 50\%, 75\%, and 100\% of its demonstrations. At 100\%, each task has 50 demonstrations: 2,000 in total across 40 LIBERO tasks or 2,500 across 50 RoboTwin tasks. As shown in Figure~\ref{fig:posttraining-data-efficiency}, Phoenix achieves 91.1\% success on LIBERO-Plus using only 25\% of the demonstrations, within 3.7 points of its full-data result of 94.8\%. With 50\% of the demonstrations, success reaches 94.0\%, approaching full-data performance.

On RoboTwin 2.0 C2R, Phoenix achieves 58.6\% success with 25\% of the demonstrations, outperforming the VAE-only Fast-WAM baseline post-trained on all task demonstrations (47.7\%; Figure~\ref{fig:pretraining-data-scaling}) by 10.9 points. Using 50\% of the demonstrations further improves success to 66.1\%, a lead of 18.4 points over this baseline. These results show that learned scene representations transfer to downstream manipulation with limited task-specific demonstrations.

\paragraph{Q4: Does Phoenix predict manipulation dynamics more faithfully?}
\begin{table}[t]
\centering\small
\caption{Video prediction quality (PT: WAM pretraining).}
\label{tab:future-prediction}
\newcommand{\futurepredictionrows}{%
\toprule
Method & {\footnotesize\mbox{PSNR\,$\uparrow$}} & {\footnotesize\mbox{SSIM\,$\uparrow$}} & {\footnotesize\mbox{LPIPS\,$\downarrow$}} & {\footnotesize\mbox{DINO-GT\,$\uparrow$}} & {\footnotesize\mbox{CLIP-GT\,$\uparrow$}} \\
\midrule
\rowcolor{benchmarkgroup}\multicolumn{6}{c}{\textit{LIBERO-Plus}} \\
Fast-WAM~\cite{fastwam} & 20.75 & 0.77 & 0.19 & 85.33 & 91.50 \\
Phoenix (w/o. PT) & 20.96 & 0.78 & 0.17 & 86.06 & 91.46 \\
\rowcolor{phoenixrow}\textbf{Phoenix (Ours)} & \textbf{23.31} & \textbf{0.81} & \textbf{0.12} & \textbf{90.28} & \textbf{93.09} \\
\midrule
\rowcolor{benchmarkgroup}\multicolumn{6}{c}{\textit{RoboTwin 2.0 Clean2Random}} \\
Fast-WAM~\cite{fastwam} & 16.51 & 0.61 & 0.39 & 60.33 & 83.27 \\
Phoenix (w/o. PT) & 21.19 & 0.75 & 0.30 & 76.02 & 88.96 \\
\rowcolor{phoenixrow}\textbf{Phoenix (Ours)} & \textbf{26.30} & \textbf{0.82} & \textbf{0.16} & \textbf{86.12} & \textbf{93.95} \\
\bottomrule
}
\setlength{\tabcolsep}{0pt}
\sbox{\benchmarkmeasure}{\begin{tabular}{lccccc}\futurepredictionrows\end{tabular}}
\setlength{\tabcolsep}{\dimexpr(\linewidth-\wd\benchmarkmeasure)/12\relax}
\typeout{FUTURE TABLE width=\the\linewidth; natural=\the\wd\benchmarkmeasure; tabcolsep=\the\tabcolsep}
\begin{tabular}{lccccc}\futurepredictionrows\end{tabular}
\end{table}

We examine whether Phoenix's manipulation generalization is accompanied by more faithful predictions of scene evolution. We evaluate 200 windows per benchmark on LIBERO-Plus and RoboTwin 2.0 C2R, matched across methods by task and window identifiers. Each model predicts eight future frames spanning 32 action steps from its own rollout observations, with only future frames scored against the corresponding ground truth. DINO-GT and CLIP-GT measure cosine similarity between predicted and ground-truth frames using frozen DINO ViT-B/16 [CLS] features and CLIP ViT-B/32 projected image embeddings, respectively; scores are averaged over future frames, views, and windows and multiplied by 100.

Table~\ref{tab:future-prediction} compares Fast-WAM with Phoenix without and with WAM pretraining. Without WAM pretraining, Phoenix achieves higher PSNR and lower LPIPS than Fast-WAM on both evaluation subsets, supporting the value of learned scene representations for future prediction. On RoboTwin 2.0 C2R, PSNR increases from 16.51 to 21.19 and LPIPS decreases from 0.39 to 0.30. WAM pretraining substantially improves these metrics to 26.30 and 0.16, respectively, with Phoenix leading all five metrics on both subsets.

Figure~\ref{fig:future-prediction} illustrates these improvements: pretrained Phoenix produces more coherent motion across future frames while preserving clearer object contours and robot structures.
These results show that Phoenix's manipulation generalization is accompanied by more faithful predictions of interaction dynamics.

\section{Conclusion}
\begingroup
\widowpenalty=10000
We presented Phoenix, which learns scene representations for generalizable world--action modeling. Phoenix Encoder builds on DINOv3's visual semantics through dual-teacher distillation of geometric and spatiotemporal semantic knowledge. CS-VAE causally aggregates and compresses Phoenix Encoder features into generative latents for joint prediction with appearance and actions. Representation learning uses approximately 39,000 hours of egocentric and robot manipulation data. After benchmark-specific post-training, Phoenix achieves SOTA performance on six simulation and real-robot benchmarks spanning perturbation robustness, dexterous manipulation, and adaptation across robot configurations. Ablations validate both scene feature learning and generative compression. Phoenix also gains more from scaling pretraining data than a video-VAE-only baseline and adapts effectively with few downstream demonstrations. Better future-video prediction complements these policy results. Together, these findings connect scene representation learning with manipulation generalization and pretraining data efficiency.
\par
\endgroup

\clearpage
\onecolumn
\appendix
\setcounter{table}{0}
\setcounter{figure}{0}
\setcounter{equation}{0}
\renewcommand{\thetable}{S\arabic{table}}
\renewcommand{\thefigure}{S\arabic{figure}}
\renewcommand{\theequation}{S\arabic{equation}}
\renewcommand{\theHtable}{supp.\arabic{table}}
\renewcommand{\theHfigure}{supp.\arabic{figure}}
\renewcommand{\theHequation}{supp.\arabic{equation}}
\renewcommand{\theHsection}{supp.\Alph{section}}
\makeatletter
\let\PhoenixOriginalAddContentsLine\addcontentsline
\renewcommand{\addcontentsline}[3]{%
  \ifstrequal{#1}{toc}
    {\PhoenixOriginalAddContentsLine{stoc}{#2}{#3}}
    {\PhoenixOriginalAddContentsLine{#1}{#2}{#3}}%
}
\makeatother
\pdfbookmark[0]{Supplementary Material}{supplementary}
\begin{center}
  {\Large\bfseries Supplementary Material\par}
  \vspace{1.2em}
  {\large\bfseries Contents\par}
\end{center}
\setcounter{tocdepth}{2}
\hypersetup{linktoc=all}
\makeatletter
\renewcommand*{\l@subsection}[2]{\@dottedtocline{2}{1.5em}{2.3em}{\hspace*{1.5em}#1}{#2}}
\@starttoc{stoc}
\makeatother
\clearpage
\twocolumn
\raggedbottom

\newsavebox{\suppTableMeasure}
\newlength{\suppTableGap}
\section{Additional Benchmark Results}
\label{sec:s-detailed-benchmarks}
\subsection{Benchmark Details}\label{sec:s-benchmark-details}
We evaluate Phoenix on six manipulation benchmarks spanning environmental perturbations, cross-embodiment adaptation, dexterous manipulation, and physical robot deployment. Below, we describe their task coverage, demonstration datasets, evaluation settings, and reporting conventions.

\paragraph{LIBERO-Plus~\cite{plus}.}
LIBERO-Plus evaluates manipulation under distribution shifts using tasks derived from the four standard LIBERO~\cite{liu2023liberobenchmarkingknowledgetransfer} suites: Spatial, Object, Goal, and Long. Each suite contains 10 tasks with 50 demonstrations per task, providing 2,000 demonstrations before preprocessing. LIBERO-Plus introduces seven perturbation categories---camera viewpoint, robot initialization, language instruction, lighting, background texture, sensor noise, and object layout---with 10,030 evaluation instances. The policy is adapted using LIBERO demonstrations and evaluated on the perturbed instances. We report success rates for each category and overall success across all 10,030 instances.

\paragraph{LIBERO-Pro~\cite{pro}.}
LIBERO-Pro evaluates policies adapted on the standard LIBERO demonstration datasets under controlled modifications to objects, spatial configurations, instructions, and task goals. Our comparison reports four perturbation settings: Object, Position, Semantic, and Task. Object perturbations change appearance, size, or color; Position perturbations relocate objects within physically valid regions; Semantic perturbations introduce alternative instruction wording; and Task perturbations modify the relevant objects or target states. We evaluate each setting on 40 tasks across the four LIBERO suites, using 50 episodes per task. We report the four setting-specific success rates and their arithmetic mean, excluding unperturbed performance from the aggregate.

\paragraph{RoboCasa-GR1~\cite{casa,gr00tn1_2025}.}
RoboCasa-GR1 comprises 24 tabletop manipulation tasks performed by the Fourier GR1 robot with two arms and dexterous hands. Eighteen tasks involve rearranging objects between source and target receptacles, while six require placing objects inside articulated fixtures, including cabinets, drawers, and microwaves, and subsequently closing them. We use 1,000 generated demonstrations per task, totaling 24,000 trajectories. Observations include an egocentric RGB view, and tasks vary object identities, placements, receptacles, and distractors. Our evaluation uses 50 rollouts per task, totaling 1,200 episodes. We report success rates for the pick-and-place and articulated-object categories, together with the equally weighted mean across all 24 tasks.

\paragraph{RoboTwin 2.0 Clean2Random~\cite{twin}.}
RoboTwin 2.0 contains 50 manipulation tasks evaluated with the ALOHA-AgileX dual-arm embodiment. The Clean2Random setting provides 50 clean expert demonstrations per task, totaling 2,500 training trajectories, and evaluates policies in randomized environments. We use 100 evaluation rollouts per task for the randomized condition, totaling 5,000 episodes. Randomization introduces scene clutter, lighting changes, background textures, tabletop-height variation, and unseen instruction wording. We report Clean2Random success averaged over the 50 tasks, assessing whether behavior learned from clean demonstrations transfers to these perturbed environments.

\paragraph{Bench2Dex~\cite{dex}.}
Bench2Dex comprises 26 task--embodiment settings spanning 12 arm--hand combinations, with approximately 1,300 human-teleoperated demonstrations. Tasks cover tool use, articulated-object interaction, and multi-stage bimanual manipulation. Evaluation uses 50 rollouts per task--embodiment setting under each of four conditions: unperturbed, equivariant perturbations, invariant perturbations, and their combination. This yields 1,300 episodes per condition and 5,200 episodes per policy. Equivariant perturbations alter object poses and table height, requiring corresponding action adjustments; invariant perturbations modify backgrounds, textures, lighting, distractors, and camera poses. We report success rate (SR) and latched stage completion rate (LSCR). For an episode $e$ with $K_e$ task stages, LSCR is defined as
\begin{equation}
\mathrm{LSCR}_e=\frac{1}{K_e}\sum_{k=1}^{K_e}L_{e,k},
\end{equation}
where $L_{e,k}=1$ if stage $k$ is reached with its dependencies satisfied by previously completed or concurrently reached stages, and $0$ otherwise. Once credited, a stage remains counted even if subsequent actions reverse its completion condition. LSCR therefore captures partial task progress beyond binary success. We average episode-level LSCR within each evaluation setting, then aggregate scores across the four conditions and 26 task--embodiment settings.

\paragraph{RoboDojo-RealWorld~\cite{dojo}.}
RoboDojo-RealWorld includes 18 tasks across three robot configurations: ARX X5, Piper, and Piper X, with six tasks per configuration. The dataset provides 100 teleoperated demonstrations per task, totaling 1,800 trajectories and approximately 17.9 hours of recordings. Demonstrations contain synchronized observations from a head camera and two wrist cameras, together with robot states and language annotations. Tasks involve long-horizon execution, precise alignment, multi-object handling, and contact-rich interaction. We conduct 10 trials per task with standardized initial layouts and scene resets, yielding 60 trials per configuration and 180 trials per policy. Success rate (SR) is the percentage of trials in which the robot completes the full task. Score measures average task progress on a 0--100 scale, giving partial credit for completed intermediate steps even when the robot does not reach the final goal. For both metrics, we average equally across the six tasks within each configuration and then across the three configurations.

\subsection{Results and Analysis}\label{sec:s-benchmark-analysis}
We extend the evaluation in Section~4.2 with two benchmark comparisons: LIBERO-Pro for generalization under perturbations (Table~\ref{tab:s-pro}) and RoboCasa-GR1 for cross-embodiment adaptation (Table~\ref{tab:s-casa}). We also report task-level results for RoboTwin 2.0 Clean2Random (Table~\ref{tab:s-twin}), Bench2Dex (Table~\ref{tab:s-dex-task}), and RoboDojo-RealWorld (Table~\ref{tab:s-dojo}). Bold and underlined values denote the best and second-best results, respectively.

\paragraph{Generalization under perturbations.}
LIBERO-Pro evaluates policies trained on standard LIBERO demonstrations under changes to object appearance, object placement, instruction wording, and task goals. Phoenix achieves the highest mean success rate across these four settings at 66.5\%, exceeding Continuous Reasoning by 2.5 percentage points (Table~\ref{tab:s-pro}). It achieves 91.3\% success under object perturbations and 98.5\% under semantic perturbations, and leads all methods under position perturbations at 42.6\%, 3.3 points above the next-best result. With scene structure modeled explicitly during WAM pretraining, Phoenix achieves the strongest overall generalization on LIBERO-Pro, with a clear advantage under changes in spatial layout.

\paragraph{Cross-embodiment adaptation.}
RoboCasa-GR1 evaluates tabletop manipulation with the Fourier GR1's two arms and dexterous hands. After benchmark-specific post-training, Phoenix achieves the highest overall success rate at 82.7\% across 24 tasks (Table~\ref{tab:s-casa}). It reaches 82.1\% on pick-and-place tasks and leads on articulated-object manipulation at 84.3\%, 4.0 points above WALA. The pretrained Phoenix model adapts to bimanual dexterous manipulation on GR1, covering both free-object transport and articulated-object interaction.

\begingroup
\renewcommand{\arraystretch}{0.88}
\noindent\begin{minipage}{\columnwidth}
\centering\small\setlength{\tabcolsep}{7pt}
\captionof{table}{LIBERO-Pro success rate.}\label{tab:s-pro}
\setlength{\tabcolsep}{0pt}
\sbox{\suppTableMeasure}{\begin{tabular}{lrrrrr}\toprule
Method & Obj. & Pos. & Sem. & Task & Avg.\\ \midrule
\rowcolor{benchmarkgroup}\multicolumn{6}{c}{\textit{VLA}}\\
OpenVLA-OFT~\cite{oft} & 27.6 & 5.8 & 59.0 & 0.6 & 23.3\\
NORA~\cite{supp_nora} & 70.5 & 0.0 & 86.3 & 0.0 & 39.0\\
MolmoAct~\cite{supp_molmo} & 76.0 & 1.5 & 85.8 & 1.5 & 41.0\\
$\pi_0$~\cite{pi0} & 90.5 & 0.0 & 90.5 & 0.0 & 45.3\\
X-VLA~\cite{xvla} & 79.0 & 0.8 & 96.8 & 6.8 & 46.0\\
VLA-Adapter~\cite{supp_adapter} & 73.8 & 0.0 & 90.8 & 19.8 & 46.1\\
OpenVLA~\cite{openvla} & \underline{93.0} & 0.0 & 97.3 & 0.0 & 47.6\\
SimVLA~\cite{supp_simvla} & 81.5 & 8.3 & \underline{98.8} & 6.0 & 48.6\\
HABILIS Brain 0~\cite{supp_habilis} & 71.9 & 18.1 & 93.0 & 21.5 & 51.1\\
$\pi_{0.5}$~\cite{pi05} & \textbf{96.0} & 20.8 & 95.8 & 0.8 & 53.3\\
Quantile Head~\cite{supp_quantile} & 85.8 & 31.3 & 97.8 & 26.0 & 60.2\\
QuoVLA~\cite{supp_quo} & 88.3 & 36.0 & \textbf{99.5} & 25.5 & 62.3\\
Continuous Reasoning~\cite{reason} & 84.6 & \underline{39.3} & 94.9 & \underline{37.1} & \underline{64.0}\\
\rowcolor{benchmarkgroup}\multicolumn{6}{c}{\textit{WAM}}\\
PatchWAM~\cite{patchwam} & 80.8 & 8.8 & 96.8 & 5.0 & 47.9\\
Flex-$\pi$~\cite{supp_flexpi} & 87.3 & 11.8 & 98.3 & 5.0 & 51.0\\
GAM~\cite{gam} & 75.8 & 32.8 & 95.0 & \textbf{40.5} & 61.0\\
\rowcolor{phoenixrow}\textbf{Phoenix (Ours)} & 91.3 & \textbf{42.6} & 98.5 & 33.6 & \textbf{66.5}\\
\bottomrule\end{tabular}}
\setlength{\suppTableGap}{\dimexpr(\linewidth-\wd\suppTableMeasure-13.5pt)/5\relax}
\typeout{SUPP pro: width=\the\linewidth; gap=\the\suppTableGap}
\begin{tabular}{p{6.75pt}lp{\suppTableGap}rp{\suppTableGap}rp{\suppTableGap}rp{\suppTableGap}rp{\suppTableGap}rp{6.75pt}}\toprule
 & Method & & Obj. & & Pos. & & Sem. & & Task & & Avg. & \\ \midrule
\rowcolor{benchmarkgroup}\multicolumn{13}{c}{\textit{VLA}}\\
 & OpenVLA-OFT~\cite{oft} & & 27.6 & & 5.8 & & 59.0 & & 0.6 & & 23.3 & \\
 & NORA~\cite{supp_nora} & & 70.5 & & 0.0 & & 86.3 & & 0.0 & & 39.0 & \\
 & MolmoAct~\cite{supp_molmo} & & 76.0 & & 1.5 & & 85.8 & & 1.5 & & 41.0 & \\
 & $\pi_0$~\cite{pi0} & & 90.5 & & 0.0 & & 90.5 & & 0.0 & & 45.3 & \\
 & X-VLA~\cite{xvla} & & 79.0 & & 0.8 & & 96.8 & & 6.8 & & 46.0 & \\
 & VLA-Adapter~\cite{supp_adapter} & & 73.8 & & 0.0 & & 90.8 & & 19.8 & & 46.1 & \\
 & OpenVLA~\cite{openvla} & & \underline{93.0} & & 0.0 & & 97.3 & & 0.0 & & 47.6 & \\
 & SimVLA~\cite{supp_simvla} & & 81.5 & & 8.3 & & \underline{98.8} & & 6.0 & & 48.6 & \\
 & HABILIS Brain 0~\cite{supp_habilis} & & 71.9 & & 18.1 & & 93.0 & & 21.5 & & 51.1 & \\
 & $\pi_{0.5}$~\cite{pi05} & & \textbf{96.0} & & 20.8 & & 95.8 & & 0.8 & & 53.3 & \\
 & Quantile Head~\cite{supp_quantile} & & 85.8 & & 31.3 & & 97.8 & & 26.0 & & 60.2 & \\
 & QuoVLA~\cite{supp_quo} & & 88.3 & & 36.0 & & \textbf{99.5} & & 25.5 & & 62.3 & \\
 & Continuous Reasoning~\cite{reason} & & 84.6 & & \underline{39.3} & & 94.9 & & \underline{37.1} & & \underline{64.0} & \\
\rowcolor{benchmarkgroup}\multicolumn{13}{c}{\textit{WAM}}\\
 & PatchWAM~\cite{patchwam} & & 80.8 & & 8.8 & & 96.8 & & 5.0 & & 47.9 & \\
 & Flex-$\pi$~\cite{supp_flexpi} & & 87.3 & & 11.8 & & 98.3 & & 5.0 & & 51.0 & \\
 & GAM~\cite{gam} & & 75.8 & & 32.8 & & 95.0 & & \textbf{40.5} & & 61.0 & \\
\rowcolor{phoenixrow} & \textbf{Phoenix (Ours)} & & 91.3 & & \textbf{42.6} & & 98.5 & & 33.6 & & \textbf{66.5} & \\
\bottomrule\end{tabular}
\end{minipage}\par

\vspace{0.75em}

\noindent\begin{minipage}{\columnwidth}
\centering\small\setlength{\tabcolsep}{7pt}
\captionof{table}{RoboCasa-GR1 success rate.}\label{tab:s-casa}
\setlength{\tabcolsep}{0pt}
\sbox{\suppTableMeasure}{\begin{tabular}{lrrr}\toprule
Method & Art. & P\&P & Avg.\\ \midrule
\rowcolor{benchmarkgroup}\multicolumn{4}{c}{\textit{VLA}}\\
$\pi_{0.5}$~\cite{pi05} & -- & -- & 37.0\\
GR00T N1.6~\cite{gr00tn1_2025} & 24.2 & 55.4 & 47.6\\
StarVLA~\cite{supp_starvla} & 43.7 & 50.4 & 48.8\\
StarVLA-$\alpha$~\cite{supp_staralpha} & 49.0 & 55.4 & 53.8\\
TwinBrainVLA~\cite{supp_twinbrain} & 56.7 & 53.9 & 54.6\\
CoRE-VLA~\cite{supp_corevla} & 56.2 & 56.6 & 56.5\\
ABot-M0~\cite{supp_abot0} & 61.7 & 57.1 & 58.3\\
RLDX-1~\cite{supp_rldx} & 57.7 & 59.0 & 58.7\\
JoyAI-RA 0.1~\cite{supp_joy} & 71.7 & 60.4 & 63.2\\
DIAL~\cite{supp_dial} & 74.3 & 68.9 & 70.2\\
ACE-Ego-0~\cite{supp_ace} & 59.0 & 77.4 & 72.8\\
\rowcolor{benchmarkgroup}\multicolumn{4}{c}{\textit{WAM}}\\
DiT4DiT~\cite{supp_dit} & 50.3 & 50.9 & 50.8\\
LDA-1B~\cite{supp_lda} & 56.3 & 55.1 & 55.4\\
OpenWAM-$\alpha$~\cite{openwam} & -- & -- & 60.5\\
Fast-WAM-IDM~\cite{fastwam} & 59.7 & 78.7 & 73.9\\
Fast-WAM~\cite{fastwam} & 62.0 & 78.8 & 74.6\\
WALA~\cite{wala} & \underline{80.3} & 73.5 & 75.2\\
LAWA (Full)~\cite{supp_lawa} & 73.0 & \underline{83.4} & 80.8\\
CF-WAM~\cite{cfwam} & 75.0 & \textbf{85.0} & \underline{82.5}\\
\rowcolor{phoenixrow}\textbf{Phoenix (Ours)} & \textbf{84.3} & 82.1 & \textbf{82.7}\\
\bottomrule\end{tabular}}
\setlength{\suppTableGap}{\dimexpr(\linewidth-\wd\suppTableMeasure-13.5pt)/3\relax}
\typeout{SUPP casa: width=\the\linewidth; gap=\the\suppTableGap}
\begin{tabular}{p{6.75pt}lp{\suppTableGap}rp{\suppTableGap}rp{\suppTableGap}rp{6.75pt}}\toprule
 & Method & & Art. & & P\&P & & Avg. & \\ \midrule
\rowcolor{benchmarkgroup}\multicolumn{9}{c}{\textit{VLA}}\\
 & $\pi_{0.5}$~\cite{pi05} & & -- & & -- & & 37.0 & \\
 & GR00T N1.6~\cite{gr00tn1_2025} & & 24.2 & & 55.4 & & 47.6 & \\
 & StarVLA~\cite{supp_starvla} & & 43.7 & & 50.4 & & 48.8 & \\
 & StarVLA-$\alpha$~\cite{supp_staralpha} & & 49.0 & & 55.4 & & 53.8 & \\
 & TwinBrainVLA~\cite{supp_twinbrain} & & 56.7 & & 53.9 & & 54.6 & \\
 & CoRE-VLA~\cite{supp_corevla} & & 56.2 & & 56.6 & & 56.5 & \\
 & ABot-M0~\cite{supp_abot0} & & 61.7 & & 57.1 & & 58.3 & \\
 & RLDX-1~\cite{supp_rldx} & & 57.7 & & 59.0 & & 58.7 & \\
 & JoyAI-RA 0.1~\cite{supp_joy} & & 71.7 & & 60.4 & & 63.2 & \\
 & DIAL~\cite{supp_dial} & & 74.3 & & 68.9 & & 70.2 & \\
 & ACE-Ego-0~\cite{supp_ace} & & 59.0 & & 77.4 & & 72.8 & \\
\rowcolor{benchmarkgroup}\multicolumn{9}{c}{\textit{WAM}}\\
 & DiT4DiT~\cite{supp_dit} & & 50.3 & & 50.9 & & 50.8 & \\
 & LDA-1B~\cite{supp_lda} & & 56.3 & & 55.1 & & 55.4 & \\
 & OpenWAM-$\alpha$~\cite{openwam} & & -- & & -- & & 60.5 & \\
 & Fast-WAM-IDM~\cite{fastwam} & & 59.7 & & 78.7 & & 73.9 & \\
 & Fast-WAM~\cite{fastwam} & & 62.0 & & 78.8 & & 74.6 & \\
 & WALA~\cite{wala} & & \underline{80.3} & & 73.5 & & 75.2 & \\
 & LAWA (Full)~\cite{supp_lawa} & & 73.0 & & \underline{83.4} & & 80.8 & \\
 & CF-WAM~\cite{cfwam} & & 75.0 & & \textbf{85.0} & & \underline{82.5} & \\
\rowcolor{phoenixrow} & \textbf{Phoenix (Ours)} & & \textbf{84.3} & & 82.1 & & \textbf{82.7} & \\
\bottomrule\end{tabular}
\end{minipage}\par

\endgroup
\onecolumn
\flushbottom

\paragraph{Task-level comparisons.}
On RoboTwin 2.0 Clean2Random, Phoenix achieves the highest or joint-highest success rate on 27 of 50 tasks, indicating that its advantage under scene randomization extends across a majority of the task set (Table~\ref{tab:s-twin}).

\begingroup
\centering\small
\setlength{\tabcolsep}{3pt}
\renewcommand{\arraystretch}{0.99}
\captionof{table}{RoboTwin 2.0 Clean2Random task-level success rates (\%).}\label{tab:s-twin}
\setlength{\aboverulesep}{0pt}\setlength{\belowrulesep}{0pt}
\begin{tabular*}{\linewidth}{@{\extracolsep{\fill}}lccccc>{\columncolor{phoenixrow}}c}\toprule
 & \multicolumn{3}{c}{\textit{VLA}} & \multicolumn{3}{c}{\textit{WAM}}\\
\cmidrule(lr){2-4}\cmidrule(lr){5-7}
Task & $\pi_0$~\cite{pi0} & $\pi_{0.5}$~\cite{pi05} & GigaBrain-0.7~\cite{supp_giga} & Fast-WAM~\cite{fastwam} & 4D-WAM~\cite{supp_fourwam} & \textbf{Phoenix (Ours)}\\\midrule
Adjust Bottle & 56.0 & 77.0 & \textbf{98.0} & 0.0 & 36.0 & \underline{85.0}\\
Beat Block Hammer & 21.0 & 23.0 & \underline{74.0} & 0.0 & 28.0 & \textbf{77.0}\\
Blocks Ranking RGB & 5.0 & 44.0 & \textbf{76.0} & 0.0 & 10.0 & \underline{65.0}\\
Blocks Ranking Size & 1.0 & 21.0 & \underline{28.0} & 0.0 & 8.0 & \textbf{37.0}\\
Click Alarmclock & 11.0 & 50.0 & \textbf{96.0} & 60.0 & 70.0 & \underline{88.0}\\
Click Bell & 3.0 & 35.0 & \underline{84.0} & 7.0 & 77.0 & \textbf{92.0}\\
Dump Bin Bigbin & 24.0 & \textbf{84.0} & \underline{66.0} & 0.0 & 57.0 & 61.0\\
Grab Roller & 80.0 & \underline{97.0} & \textbf{100.0} & 2.0 & 68.0 & 88.0\\
Handover Block & 8.0 & 12.0 & \underline{32.0} & 0.0 & 25.0 & \textbf{65.0}\\
Handover Mic & 13.0 & 6.0 & \textbf{100.0} & 0.0 & 13.0 & \underline{89.0}\\
Hanging Mug & 3.0 & 14.0 & \underline{36.0} & 0.0 & 10.0 & \textbf{50.0}\\
Lift Pot & 36.0 & 35.0 & \textbf{86.0} & 0.0 & 9.0 & \underline{47.0}\\
Move Can Pot & 21.0 & 10.0 & \textbf{82.0} & 0.0 & 17.0 & \underline{76.0}\\
Move Pillbottle Pad & 1.0 & 32.0 & \underline{58.0} & 0.0 & 22.0 & \textbf{76.0}\\
Move Playingcard Away & 22.0 & 65.0 & \underline{82.0} & 0.0 & 35.0 & \textbf{93.0}\\
Move Stapler Pad & 2.0 & 18.0 & \underline{40.0} & 0.0 & 12.0 & \textbf{49.0}\\
Open Laptop & 46.0 & 68.0 & \textbf{88.0} & 0.0 & \underline{82.0} & 67.0\\
Open Microwave & \underline{50.0} & \textbf{59.0} & \underline{50.0} & 0.0 & 12.0 & 35.0\\
Pick Diverse Bottles & 6.0 & 42.0 & \underline{50.0} & 0.0 & 44.0 & \textbf{76.0}\\
Pick Dual Bottles & 12.0 & 57.0 & 56.0 & 1.0 & \underline{67.0} & \textbf{94.0}\\
Place A2B Left & 1.0 & 49.0 & \textbf{72.0} & 0.0 & 19.0 & \underline{68.0}\\
Place A2B Right & 6.0 & 39.0 & \underline{68.0} & 1.0 & 42.0 & \textbf{74.0}\\
Place Bread Basket & 4.0 & 60.0 & 68.0 & 0.0 & \textbf{77.0} & \underline{76.0}\\
Place Bread Skillet & 1.0 & 47.0 & 50.0 & 0.0 & \textbf{69.0} & \underline{51.0}\\
Place Burger Fries & 4.0 & \textbf{86.0} & \textbf{86.0} & 0.0 & 72.0 & \underline{73.0}\\
Place Can Basket & 5.0 & 10.0 & 22.0 & 0.0 & \underline{51.0} & \textbf{56.0}\\
Place Cans Plasticbox & 2.0 & \underline{73.0} & 18.0 & 0.0 & 48.0 & \textbf{87.0}\\
Place Container Plate & 45.0 & 72.0 & \textbf{94.0} & 0.0 & 82.0 & \underline{89.0}\\
Place Dual Shoes & 0.0 & 45.0 & \textbf{72.0} & 0.0 & 19.0 & \underline{52.0}\\
Place Empty Cup & 11.0 & 80.0 & \textbf{90.0} & 0.0 & 70.0 & \underline{83.0}\\
Place Fan & 10.0 & 26.0 & \underline{74.0} & 0.0 & 17.0 & \textbf{77.0}\\
Place Mouse Pad & 1.0 & 31.0 & \underline{44.0} & 0.0 & 9.0 & \textbf{61.0}\\
Place Object Basket & 2.0 & 29.0 & \underline{64.0} & 0.0 & 40.0 & \textbf{83.0}\\
Place Object Scale & 0.0 & 38.0 & \underline{54.0} & 0.0 & 24.0 & \textbf{77.0}\\
Place Object Stand & 11.0 & 47.0 & \underline{76.0} & 0.0 & 42.0 & \textbf{87.0}\\
Place Phone Stand & 7.0 & 37.0 & \underline{52.0} & 0.0 & 21.0 & \textbf{82.0}\\
Place Shoe & 6.0 & 50.0 & \textbf{92.0} & 0.0 & 52.0 & \underline{90.0}\\
Press Stapler & 29.0 & 63.0 & \underline{74.0} & 3.0 & 61.0 & \textbf{85.0}\\
Put Bottles Dustbin & 13.0 & 50.0 & \underline{54.0} & 0.0 & 19.0 & \textbf{61.0}\\
Put Object Cabinet & 18.0 & 22.0 & \textbf{82.0} & 0.0 & \underline{47.0} & 41.0\\
Rotate QRcode & 15.0 & 21.0 & \underline{58.0} & 0.0 & \textbf{67.0} & 54.0\\
Scan Object & 1.0 & 40.0 & 46.0 & 0.0 & \underline{53.0} & \textbf{72.0}\\
Shake Bottle Horizontally & 51.0 & \textbf{100.0} & \underline{92.0} & 9.0 & 70.0 & \textbf{100.0}\\
Shake Bottle & 60.0 & \textbf{100.0} & \underline{98.0} & 10.0 & 72.0 & \textbf{100.0}\\
Stack Blocks Three & 0.0 & 25.0 & \underline{78.0} & 0.0 & \textbf{82.0} & 73.0\\
Stack Blocks Two & 1.0 & 43.0 & \textbf{92.0} & 0.0 & 34.0 & \underline{88.0}\\
Stack Bowls Three & 24.0 & 52.0 & \underline{64.0} & 0.0 & 27.0 & \textbf{76.0}\\
Stack Bowls Two & 41.0 & 70.0 & 74.0 & 0.0 & \underline{81.0} & \textbf{85.0}\\
Stamp Seal & 4.0 & 21.0 & \underline{54.0} & 0.0 & 10.0 & \textbf{60.0}\\
Turn Switch & 23.0 & 25.0 & \textbf{52.0} & 3.0 & 11.0 & \underline{50.0}\\
\midrule
Task mean (50 tasks) & 16.3 & 46.0 & \underline{67.9} & 1.9 & 41.8 & \textbf{72.4}\\
\bottomrule\end{tabular*}\par
\smallskip
\endgroup
\clearpage

\noindent On Bench2Dex, Phoenix leads or ties for the lead in success rate on 21 of 26 task--embodiment settings. It also exceeds GR00T N1.5 in mean latched stage completion rate (65.6\% vs.\ 47.3\%), showing gains in both intermediate-stage completion and end-to-end task success (Table~\ref{tab:s-dex-task}).

\begingroup
\begingroup
\centering\footnotesize\setlength{\tabcolsep}{2pt}
\renewcommand{\arraystretch}{1.3}
\setlength{\aboverulesep}{0pt}\setlength{\belowrulesep}{0pt}
\captionof{table}{Bench2Dex task-level success rate (SR) and latched stage completion rate (LSCR).}\label{tab:s-dex-task}
\begin{adjustbox}{width=\linewidth}
\begin{tabular}{p{116pt}>{\centering\arraybackslash}p{90pt}*{8}{>{\centering\arraybackslash}p{27pt}}*{2}{>{\columncolor{phoenixrow}\centering\arraybackslash}p{27pt}}}\toprule
 & & \multicolumn{4}{c}{\textit{Visuomotor Policy}} & \multicolumn{4}{c}{\textit{VLA}} & \multicolumn{2}{c}{\textit{WAM}}\\
\cmidrule(lr){3-6}\cmidrule(lr){7-10}\cmidrule(lr){11-12}
Task & Embodiment & \multicolumn{2}{c}{ACT~\cite{act}} & \multicolumn{2}{c}{DP~\cite{dp}} & \multicolumn{2}{c}{$\pi_{0.5}$~\cite{pi05}} & \multicolumn{2}{c}{\makebox[0pt][c]{GR00T N1.5~\cite{gr00tn1_2025}}} & \multicolumn{2}{>{\columncolor{phoenixrow}}c}{\textbf{Phoenix (Ours)}}\\
 & & SR & LSCR & SR & LSCR & SR & LSCR & SR & LSCR & SR & LSCR\\ \midrule
Baking Tray Prep & \multirow{3}{90pt}{\centering IIWA7 + Sharpa} & 6.0 & 11.6 & 2.0 & 7.9 & \underline{27.5} & \underline{38.4} & 10.0 & 29.3 & \textbf{30.0} & \textbf{61.8}\\
Canned Food Tray Arrangement &  & 9.5 & 34.6 & 0.0 & 14.8 & 12.0 & \underline{38.8} & \underline{12.5} & 37.3 & \textbf{13.5} & \textbf{52.6}\\
Jigsaw Puzzle Assembly &  & \underline{0.0} & 12.5 & \underline{0.0} & 6.3 & \underline{0.0} & 8.4 & \textbf{5.0} & \textbf{30.3} & \underline{0.0} & \underline{27.6}\\
\arrayrulecolor{black!30}\hline\arrayrulecolor{black}
Gaming Desk Setup & \multirow{1}{90pt}{\centering JAKA ZU7 + DexHand021} & 31.0 & 46.0 & 9.0 & 30.7 & 33.5 & 49.3 & \textbf{59.5} & \underline{72.2} & \underline{52.0} & \textbf{73.0}\\
\arrayrulecolor{black!30}\hline\arrayrulecolor{black}
Medicine Shoebox Pack & \multirow{2}{90pt}{\centering Panda + Orca} & 7.5 & 21.5 & 0.0 & 9.3 & \underline{21.5} & \underline{39.8} & 13.0 & 26.2 & \textbf{49.5} & \textbf{65.5}\\
Tool Box Loading &  & 20.0 & 37.5 & 16.5 & 39.8 & \textbf{72.0} & \textbf{86.0} & 40.0 & 51.0 & \underline{68.5} & \underline{84.2}\\
\arrayrulecolor{black!30}\hline\arrayrulecolor{black}
Sports Ball Cup Sort & \multirow{1}{90pt}{\centering Panda + Allegro} & 2.0 & 2.5 & 0.5 & 4.4 & \underline{21.5} & \underline{33.6} & 9.0 & 21.5 & \textbf{23.0} & \textbf{43.1}\\
\arrayrulecolor{black!30}\hline\arrayrulecolor{black}
Faucet Cup Water Fill & \multirow{2}{90pt}{\centering RM65 + Revo2} & 7.5 & 35.4 & 4.0 & 14.0 & 15.5 & 40.5 & \underline{33.0} & \underline{55.0} & \textbf{35.5} & \textbf{66.5}\\
Stationery Category Sorting &  & 1.5 & \underline{23.1} & 2.0 & 18.5 & 0.0 & 2.3 & \underline{3.0} & 16.9 & \textbf{24.5} & \textbf{55.5}\\
\arrayrulecolor{black!30}\hline\arrayrulecolor{black}
Bimanual Piano Melody & \multirow{2}{90pt}{\centering xArm7 + Ability} & 4.0 & 4.5 & 0.5 & 0.5 & \underline{7.5} & \underline{16.0} & \textbf{23.0} & \textbf{23.5} & 4.0 & 6.2\\
Wine Glass Plate Balance &  & 26.0 & 72.0 & 3.5 & 39.8 & \underline{35.0} & \underline{76.2} & 28.5 & 72.5 & \textbf{40.0} & \textbf{81.8}\\
\arrayrulecolor{black!30}\hline\arrayrulecolor{black}
Cleaner Box Loading & \multirow{3}{90pt}{\centering xArm7 + LEAP} & 33.0 & 76.2 & 9.0 & 32.8 & \underline{55.0} & \underline{88.5} & 47.0 & 82.3 & \textbf{72.5} & \textbf{95.8}\\
Shoebox Accessory Pack &  & 30.0 & 56.3 & 9.5 & 35.2 & \underline{51.0} & \underline{68.3} & 40.5 & 60.3 & \textbf{64.5} & \textbf{79.8}\\
Toilet Lid Cleaner Pour &  & 34.0 & 50.3 & 8.5 & 23.5 & \underline{45.5} & 53.3 & 45.0 & \underline{63.3} & \textbf{74.5} & \textbf{84.2}\\
\arrayrulecolor{black!30}\hline\arrayrulecolor{black}
Breadbasket Fast-Food Loading & \multirow{3}{90pt}{\centering UR5 + RH5DG2} & 10.5 & \underline{15.8} & 4.5 & 7.3 & \underline{12.5} & 15.6 & 9.0 & 13.5 & \textbf{65.5} & \textbf{71.9}\\
Citrus Plate Loading &  & 51.5 & 69.9 & 22.5 & 44.4 & 56.5 & 67.3 & \underline{59.0} & \underline{73.1} & \textbf{92.5} & \textbf{96.1}\\
Fridge Wine Interhand Pour &  & 11.0 & 48.5 & 2.5 & 41.8 & 8.5 & 36.1 & \underline{21.0} & \underline{58.6} & \textbf{64.5} & \textbf{82.1}\\
\arrayrulecolor{black!30}\hline\arrayrulecolor{black}
Fruit Bowl Loading & \multirow{3}{90pt}{\centering UR5 + RH56DFX} & 37.0 & 62.0 & 18.0 & 47.8 & \underline{38.5} & \underline{71.6} & 32.5 & 68.1 & \textbf{60.0} & \textbf{82.4}\\
Screwdriver Box \& Hammer &  & 24.5 & 56.2 & 20.5 & 49.0 & 33.0 & 57.5 & \underline{48.0} & \underline{73.7} & \textbf{64.0} & \textbf{85.9}\\
Trash Disposal &  & 18.5 & 30.3 & 14.0 & 21.5 & \underline{41.0} & \underline{53.7} & 34.5 & 47.8 & \textbf{44.5} & \textbf{64.8}\\
\arrayrulecolor{black!30}\hline\arrayrulecolor{black}
Fridge Fruit Shelf Sorting & \multirow{2}{90pt}{\centering UR5 + Shadow} & 0.5 & 2.6 & 0.0 & 15.1 & \underline{6.0} & \underline{29.6} & 1.0 & 19.6 & \textbf{30.0} & \textbf{49.8}\\
Soup Serving &  & \underline{1.0} & 11.5 & 0.0 & 3.1 & 0.0 & \underline{14.0} & \textbf{3.0} & 13.9 & \textbf{3.0} & \textbf{18.8}\\
\arrayrulecolor{black!30}\hline\arrayrulecolor{black}
Frypan Stand \& Pour & \multirow{2}{90pt}{\centering UR5 + Schunk} & 19.0 & 51.2 & 14.5 & 30.0 & 15.0 & 33.3 & \underline{54.5} & \underline{72.3} & \textbf{60.0} & \textbf{85.8}\\
Microwave Bowl Loading &  & 62.5 & 74.9 & 21.0 & 48.6 & 70.0 & 78.4 & \underline{70.5} & \underline{82.0} & \textbf{90.5} & \textbf{94.1}\\
\arrayrulecolor{black!30}\hline\arrayrulecolor{black}
Ball Box Loading & \multirow{2}{90pt}{\centering UR5 + Wuji} & 8.5 & 10.4 & 0.0 & 7.1 & \underline{14.0} & 19.4 & \textbf{23.0} & \textbf{37.1} & 9.5 & \underline{19.9}\\
Condiment Box Loading &  & 10.5 & 31.8 & 1.0 & 4.2 & \underline{15.0} & \underline{33.2} & 10.5 & 27.7 & \textbf{49.5} & \textbf{76.7}\\
\midrule
\multicolumn{2}{l}{Task mean (26 tasks)} & 18.0 & 36.5 & 7.1 & 23.0 & 27.2 & 44.2 & \underline{28.3} & \underline{47.3} & \textbf{45.6} & \textbf{65.6}\\
\bottomrule\end{tabular}\end{adjustbox}\par
\endgroup
\par\endgroup

\clearpage

\noindent On RoboDojo-RealWorld, Phoenix leads in both mean success rate and score on ARX X5 and Piper X. Its largest task-level success margin over OpenWAM-$\alpha$ is on packing objects into a backpack with Piper X, where Phoenix succeeds in 70.0\% of trials versus 20.0\% (Table~\ref{tab:s-dojo}).

\smallskip\noindent\begin{minipage}{\linewidth}
\centering\footnotesize\setlength{\tabcolsep}{3pt}
\renewcommand{\arraystretch}{1.23}
\setlength{\aboverulesep}{0pt}\setlength{\belowrulesep}{0pt}
\captionof{table}{RoboDojo-RealWorld task-level success rate (SR, \%) and score (0--100).}\label{tab:s-dojo}
\begin{adjustbox}{width=\linewidth}
\begin{tabular}{lc*{10}{c}*{2}{>{\columncolor{phoenixrow}}c}}\toprule
 & & \multicolumn{8}{c}{\textit{VLA}} & \multicolumn{4}{c}{\textit{WAM}}\\
\cmidrule(lr){3-10}\cmidrule(lr){11-14}
Task & Embodiment & \multicolumn{2}{c}{StarVLA~\cite{supp_starvla}} & \multicolumn{2}{c}{$\pi_0$~\cite{pi0}} & \multicolumn{2}{c}{X-VLA~\cite{xvla}} & \multicolumn{2}{c}{$\pi_{0.5}$~\cite{pi05}} & \multicolumn{2}{c}{OpenWAM-$\alpha$~\cite{openwam}} & \multicolumn{2}{>{\columncolor{phoenixrow}}c}{\textbf{Phoenix (Ours)}}\\
 & & SR & Score & SR & Score & SR & Score & SR & Score & SR & Score & SR & Score\\ \midrule
Cover blocks & \multirow{6}{*}{ARX X5} & 0.0 & 0.0 & 0.0 & 9.0 & 0.0 & 0.0 & \underline{20.0} & 24.6 & 0.0 & \underline{31.0} & \textbf{30.0} & \textbf{46.0}\\
Store in safe &  & 0.0 & 0.0 & 0.0 & 0.0 & 10.0 & 20.7 & \underline{20.0} & \underline{40.0} & \textbf{100.0} & \textbf{100.0} & \textbf{100.0} & \textbf{100.0}\\
Pack and pour fruit &  & 0.0 & 0.0 & 0.0 & 18.0 & 0.0 & 2.0 & \underline{20.0} & 47.0 & \underline{20.0} & \underline{52.0} & \textbf{40.0} & \textbf{62.0}\\
Make bread &  & \textbf{0.0} & \underline{0.0} & \textbf{0.0} & \underline{0.0} & \textbf{0.0} & \underline{0.0} & \textbf{0.0} & \textbf{1.8} & \textbf{0.0} & \underline{0.0} & \textbf{0.0} & \underline{0.0}\\
Make food &  & 0.0 & 0.0 & 0.0 & 3.0 & 0.0 & 0.0 & \underline{10.0} & \underline{25.8} & 0.0 & 13.0 & \textbf{20.0} & \textbf{44.0}\\
Insert tubes &  & 0.0 & 0.0 & 0.0 & 18.7 & 0.0 & 18.0 & \underline{10.0} & 26.8 & \textbf{20.0} & \textbf{38.0} & \underline{10.0} & \underline{36.0}\\
\textit{Mean (ARX X5)} & & 0.0 & 0.0 & 0.0 & 8.1 & 1.7 & 6.8 & 13.3 & 27.7 & \underline{23.3} & \underline{39.0} & \textbf{33.3} & \textbf{48.0}\\
\arrayrulecolor{black!30}\hline\arrayrulecolor{black}
Stand up bottles & \multirow{6}{*}{Piper} & 0.0 & 18.7 & 0.0 & 6.0 & 0.0 & 37.0 & \textbf{50.0} & \underline{72.0} & \textbf{50.0} & \textbf{75.0} & \underline{30.0} & 65.0\\
Stack bowls &  & 30.0 & 39.0 & 30.0 & 42.0 & 50.0 & 53.0 & \underline{60.0} & \underline{72.0} & \textbf{100.0} & \textbf{100.0} & \textbf{100.0} & \textbf{100.0}\\
Stack and cover blocks &  & \underline{0.0} & 0.0 & \underline{0.0} & 0.0 & \underline{0.0} & 0.0 & \textbf{10.0} & \underline{10.0} & \underline{0.0} & 8.3 & \underline{0.0} & \textbf{20.0}\\
Insert charger &  & \textbf{0.0} & \textbf{0.0} & \textbf{0.0} & \textbf{0.0} & \textbf{0.0} & \textbf{0.0} & \textbf{0.0} & \textbf{0.0} & \textbf{0.0} & \textbf{0.0} & \textbf{0.0} & \textbf{0.0}\\
Fill pen holder &  & \underline{0.0} & 14.0 & \underline{0.0} & 6.0 & \underline{0.0} & 5.3 & \underline{0.0} & 28.0 & \textbf{40.0} & \textbf{60.0} & \underline{0.0} & \underline{30.0}\\
Put objects into basket &  & 0.0 & 0.0 & 0.0 & 0.0 & 0.0 & 0.0 & 10.0 & 10.0 & \underline{30.0} & \underline{36.7} & \textbf{60.0} & \textbf{60.0}\\
\textit{Mean (Piper)} & & 5.0 & 12.0 & 5.0 & 9.0 & 8.3 & 15.9 & 21.7 & 32.0 & \textbf{36.7} & \textbf{46.7} & \underline{31.7} & \underline{45.8}\\
\arrayrulecolor{black!30}\hline\arrayrulecolor{black}
Sweep blocks & \multirow{6}{*}{Piper X} & \underline{0.0} & 0.0 & \underline{0.0} & 0.0 & \underline{0.0} & 0.0 & \underline{0.0} & \underline{7.5} & \textbf{40.0} & \textbf{40.0} & \textbf{40.0} & \textbf{40.0}\\
Disassemble LEGO &  & \underline{0.0} & 2.7 & \underline{0.0} & 0.0 & \underline{0.0} & 0.0 & \textbf{10.0} & \underline{14.8} & \textbf{10.0} & \textbf{28.0} & \underline{0.0} & 8.0\\
Classify objects &  & \underline{0.0} & \underline{0.0} & \underline{0.0} & \underline{0.0} & \underline{0.0} & \underline{0.0} & \underline{0.0} & \underline{0.0} & \underline{0.0} & \underline{0.0} & \textbf{40.0} & \textbf{52.0}\\
Hang mugs &  & \textbf{0.0} & 0.0 & \textbf{0.0} & 0.0 & \textbf{0.0} & 0.0 & \textbf{0.0} & 0.0 & \textbf{0.0} & \textbf{18.0} & \textbf{0.0} & \underline{2.0}\\
Cap pen &  & \underline{0.0} & 0.0 & \underline{0.0} & 0.0 & \underline{0.0} & 0.7 & \underline{0.0} & 3.0 & \textbf{10.0} & \textbf{24.7} & \underline{0.0} & \underline{14.0}\\
Pack objects into backpack &  & 0.0 & 0.0 & 0.0 & 2.0 & 0.0 & 0.0 & 10.0 & 29.5 & \underline{20.0} & \underline{52.3} & \textbf{70.0} & \textbf{70.0}\\
\textit{Mean (Piper X)} & & 0.0 & 0.5 & 0.0 & 0.3 & 0.0 & 0.1 & 3.3 & 9.1 & \underline{13.3} & \underline{27.2} & \textbf{25.0} & \textbf{31.0}\\
\midrule
Task mean (18 tasks) & & 1.7 & 4.1 & 1.7 & 5.8 & 3.3 & 7.6 & 12.8 & 22.9 & \underline{24.4} & \underline{37.6} & \textbf{30.0} & \textbf{41.6}\\
\bottomrule\end{tabular}
\end{adjustbox}
\end{minipage}\par

\clearpage
\twocolumn
\section{Additional Probing Results}\label{sec:s-evaluation}
\subsection{Experimental Setup}
\label{sec:s-probing}

We use five probes from Yin et al.~\cite{yin2026probingspatialstaterepresentations} to assess spatial understanding and motion-conditioned prediction in frozen encoder representations. The evaluation uses camera trajectories with known poses from the ScanNet++ subset of InsScene-15K~\cite{yeshwanthliu2023scannetpp,li2025iggt}, comprising 2,445 training windows from 612 scenes and 276 validation windows from 69 disjoint scenes. Each window contains 28 observations sampled approximately uniformly in cumulative camera motion, measured as translation distance plus 0.5 times the rotation angle in radians, with a sampling step of 0.35. Windows are constructed with a stride of eight sampled observations, additionally including the final valid start. When more than four windows are available, four evenly spaced starts are retained per scene. We independently encode observation prefixes of $p\in\{8,12,16,24\}$ frames. For predictive readouts, we evaluate future targets at horizons $h\in\{1,2,4\}$ after the last observed frame. These horizons count sampled observations. A separate probe is trained for each encoder, task, and prefix length, and the reported scores are averaged over the four prefix lengths.

We follow the task-specific readout designs and training schedules of Yin et al., adapting the input dimensions to each encoder. For Phoenix, we use the frozen exponential-moving-average encoder employed downstream. Each $256\times256$ input frame yields a $16\times16$ grid of 1,024-dimensional patch features; the CLS token is excluded. Prefix features are extracted and cached independently, and future targets are obtained through separate encoding of the target frames. All Phoenix probes are trained on one GPU with a batch size of two, BF16 mixed precision, and seed 42. We use AdamW with $(\beta_1,\beta_2)=(0.9,0.95)$ and weight decay 0.05. Depth and instance-identity probes are trained for 100 epochs at a learning rate of $10^{-4}$, with ten warmup epochs; the remaining probes are trained for 50 epochs at $3\times10^{-4}$, with five warmup epochs. Both schedules use cosine decay to $10^{-6}$. Evaluation uses the final-epoch checkpoint.

\textit{Spatial readouts.} Depth and instance identity are decoded using pixel-aligned DPT-style heads with a trainable four-layer feature aggregator, supervising every frame in the observed prefix. The depth probe predicts metric depth and includes the camera-pose auxiliary supervision used in the reference implementation, with unit weights for the depth and camera losses. We report the mean validity-weighted absolute depth error. The instance-identity probe produces 63-dimensional pixel embeddings. For each instance, a prototype is computed by averaging its pixel embeddings across the prefix, using the ground-truth instance groups. Cross-entropy supervision encourages each pixel to match its corresponding prototype. The reported identity loss measures separation between instances across the observed views.

The view-consistency probe spatially pools each frame's features and uses a multilayer perceptron to predict overlap between pairs of prefix frames. Ground-truth overlap is computed by projecting valid depth pixels from one view into the other and taking the minimum of the two directional overlap fractions. Pairs with overlap of at least 0.40 are positive, those with overlap of at most 0.05 are negative, and intermediate pairs are excluded. A predicted probability above 0.5 indicates overlap. We report balanced accuracy, computed as the mean of the positive- and negative-class accuracies.

\textit{Predictive readouts.} Direct prediction estimates the spatially pooled feature of a future frame from the observed prefix and the relative camera motion from its final frame to the target. Camera motion is represented by a nine-dimensional vector containing a six-dimensional rotation representation and three-dimensional translation. Spatially pooled frame features and the motion input are processed by a two-layer, eight-head Transformer with hidden dimension 512. Training minimizes the sum of mean squared error and cosine distance to the target feature. Each training prefix is assigned one horizon through a fixed seeded partition, yielding 815 examples for each of $h=1,2,4$. At evaluation, every validation prefix is evaluated at all three horizons.

Path integration instead initializes a two-layer GRU with hidden dimension 512 from features pooled over the observed prefix's temporal and spatial dimensions. It sequentially applies relative camera motions from the final observation to $h=1$, from $h=1$ to $h=2$, and from $h=2$ to $h=4$, predicting a future feature at each waypoint. The path-integration probe is supervised at all three waypoints using the same MSE and cosine-distance objective. Both predictive readouts are evaluated by Global R@1 within each encoder's feature space. Each predicted feature is ranked by cosine similarity against a gallery containing the 828 target features from all 276 validation windows and three horizons. A prediction is counted as correct when its associated target ranks first.

\textit{Visualization conventions.} All qualitative examples use eight-frame observation prefixes and outputs from the trained probes. The depth comparisons (Fig.~\ref{fig:s-probing-depth}) show the input, ground truth, and predicted metric depth using a shared 0--6\,m color scale; larger depths are clipped for display, and black pixels denote missing ground truth. For instance identity (Fig.~\ref{fig:s-probing-identity}), PCA is fitted jointly to the pixel embeddings from the two displayed views of each window. PCA colors encode embedding variation; ground-truth overlays use separate colors to identify instances. View-consistency examples (Fig.~\ref{fig:s-probing-view}) display the input pair, geometric overlap, and predicted overlap probability under the thresholds defined above; example F is a false negative.

For predictive readouts, the displayed retrievals are real validation frames selected using predicted features. Path-integration examples (Fig.~\ref{fig:s-probing-path}) place the true future frames above their top-ranked retrieved candidates at $h=1,2,4$. Direct-prediction examples (Fig.~\ref{fig:s-probing-direct}) show the last observed frame, true future frame, and top-ranked retrieval. The displayed rank always refers to the true target's position among all 828 candidates. The examples include both successful and unsuccessful top-1 retrievals across the three prediction horizons.

\subsection{Qualitative Probing Results}
\label{sec:s-spatial-probing-qualitative}

The qualitative probes examine two properties of the frozen Phoenix Encoder: what scene structure its representation preserves in the current observation, and whether that representation distinguishes the view reached after a camera motion. This distinction matters for prediction: two views can share much of the same scene yet differ in which surfaces are visible and how they appear relative to one another.

\begin{figure}[!htbp]
\centering
\includegraphics[trim=0 0 7.9bp 0,clip,width=\textwidth,height=.76\textheight,keepaspectratio]{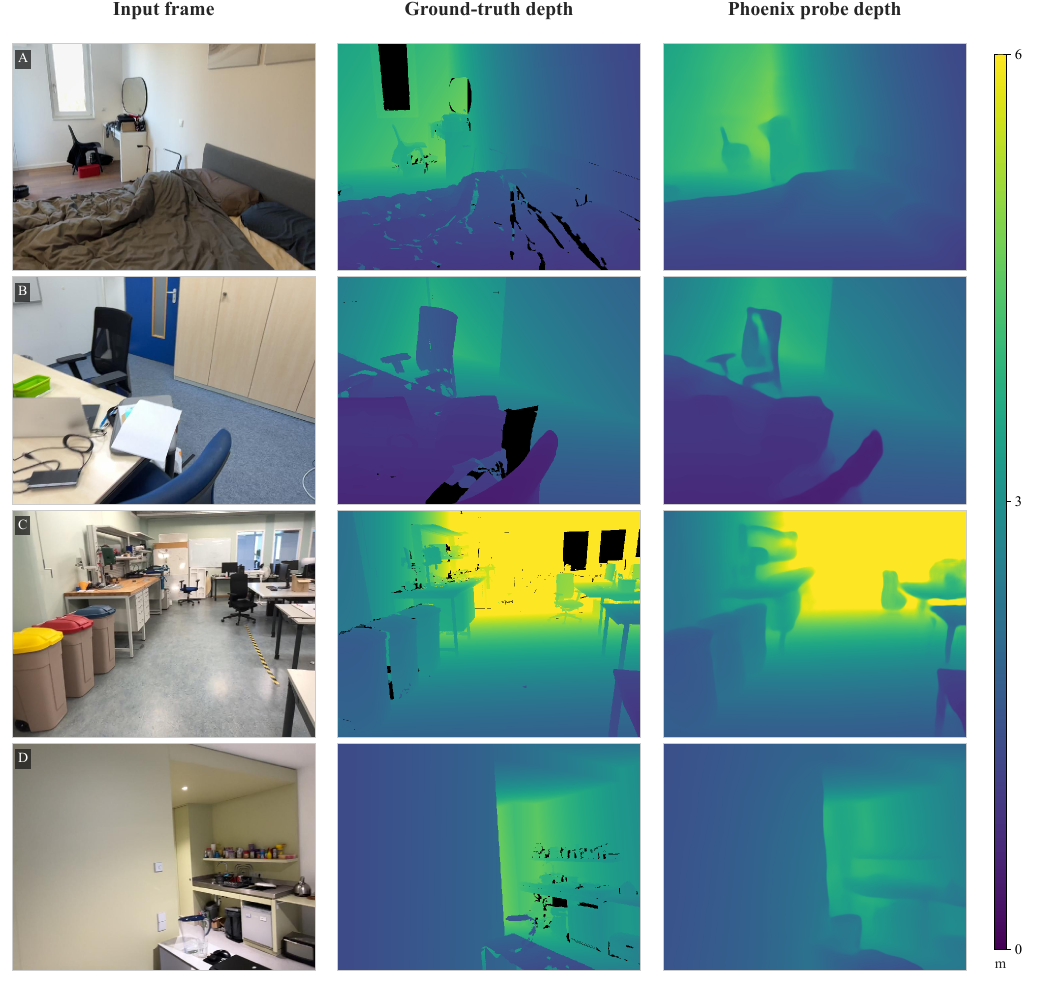}
\caption{Metric-depth probing. The Phoenix Encoder preserves both gradual changes in depth across a room and sharp depth discontinuities between adjacent surfaces.}
\label{fig:s-probing-depth}
\end{figure}

\begin{figure}[!htbp]
\centering
\includegraphics[trim=0 0 34.3bp 0,clip,width=\textwidth,height=.76\textheight,keepaspectratio]{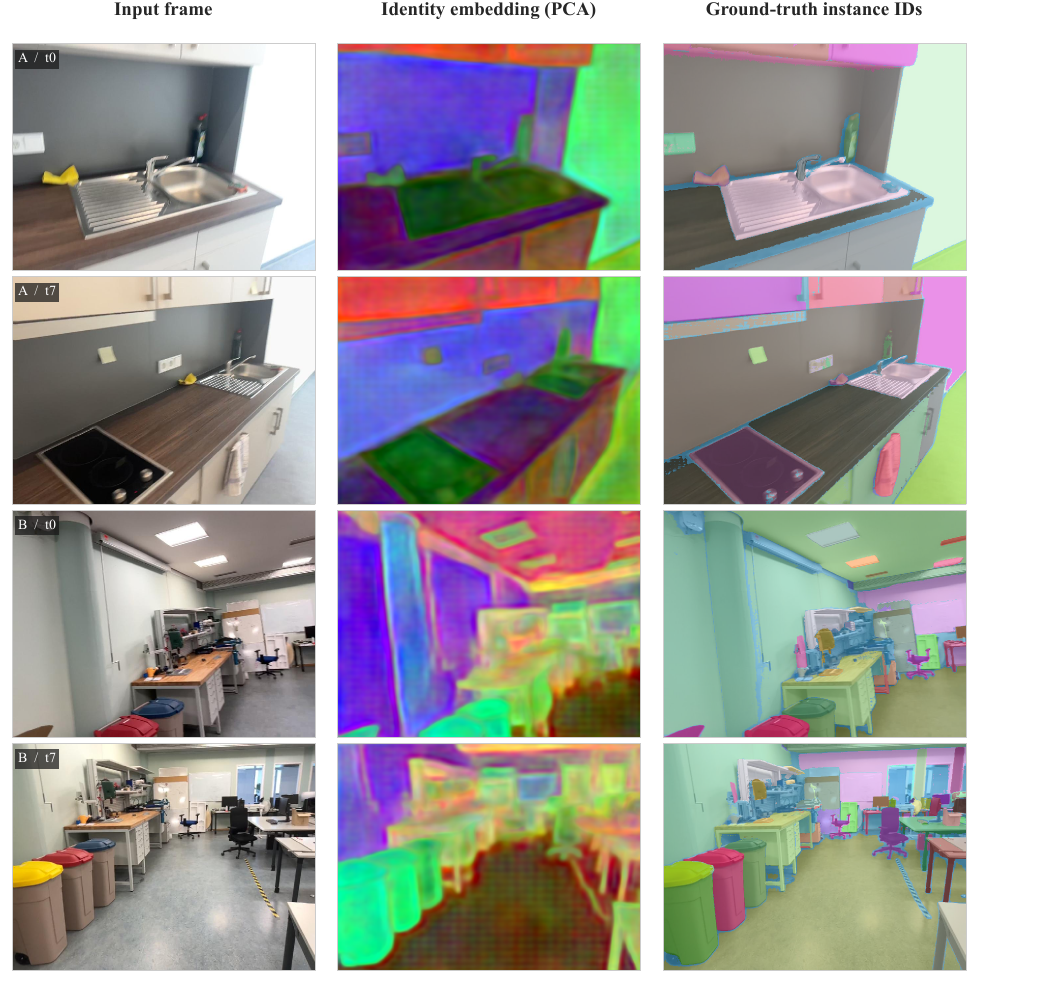}
\caption{Instance-identity probing. Object boundaries remain distinct in the Phoenix representation as objects change position and apparent scale in the image.}
\label{fig:s-probing-identity}
\end{figure}

\textit{Scene geometry and object boundaries.} The Phoenix Encoder preserves scene structure at both room and object scales. In the laboratory, decoded depth increases along the aisle from the foreground bins, past the workbenches, to the rear wall (Fig.~\ref{fig:s-probing-depth}C). In the kitchenette, the edge of the foreground wall marks an abrupt transition to the recessed counter behind it (Fig.~\ref{fig:s-probing-depth}D). Within the counter, however, the sink and draining board occupy a region close in depth to the surrounding work surface. This region is distinct in the instance-embedding map and remains discernible when the camera pulls back and the sink appears smaller near the image edge (Fig.~\ref{fig:s-probing-identity}A). The same frozen representation thus preserves both the depth relationship between surfaces and object boundaries that remain visible across viewpoints.

\begin{figure}[!htbp]
\centering
\includegraphics[trim=0 0 11.3bp 0,clip,width=\textwidth,height=.34\textheight,keepaspectratio]{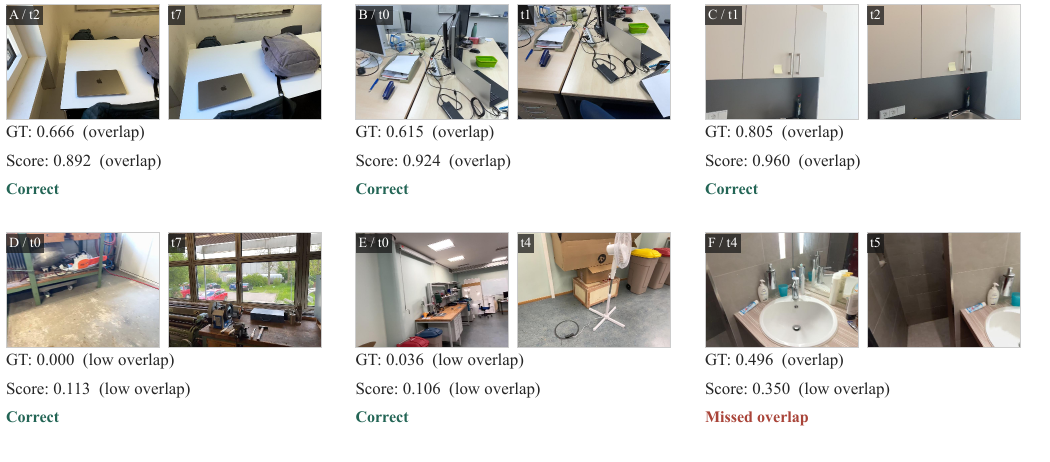}
\caption{View-overlap classification. Views of the same workspace remain recognizable as overlapping even when shared objects appear at different image locations.}
\label{fig:s-probing-view}
\end{figure}

\textit{Shared structure across viewpoints.} The two desk pairs illustrate how the Phoenix representation encodes overlap as framing changes. In Fig.~\ref{fig:s-probing-view}A, the second view includes more of the tabletop and places the laptop at a different image location. In Fig.~\ref{fig:s-probing-view}B, the monitor and nearby objects remain visible within two differently framed views of a cluttered desk. Both pairs are correctly identified as overlapping, while the pairs with little common content in Fig.~\ref{fig:s-probing-view}D--E are classified as low overlap. The basin pair in Fig.~\ref{fig:s-probing-view}F is a missed overlap: the basin is prominent in the first view but largely outside the second, and the predicted overlap probability is 0.350 for a geometric overlap of 0.496. In the overlapping desk views, shared scene content remains identifiable in the Phoenix representation even as its position in the image changes.

\begin{figure}[!htbp]
\centering
\includegraphics[trim=0 0 4.1bp 0,clip,width=\textwidth,height=.49\textheight,keepaspectratio]{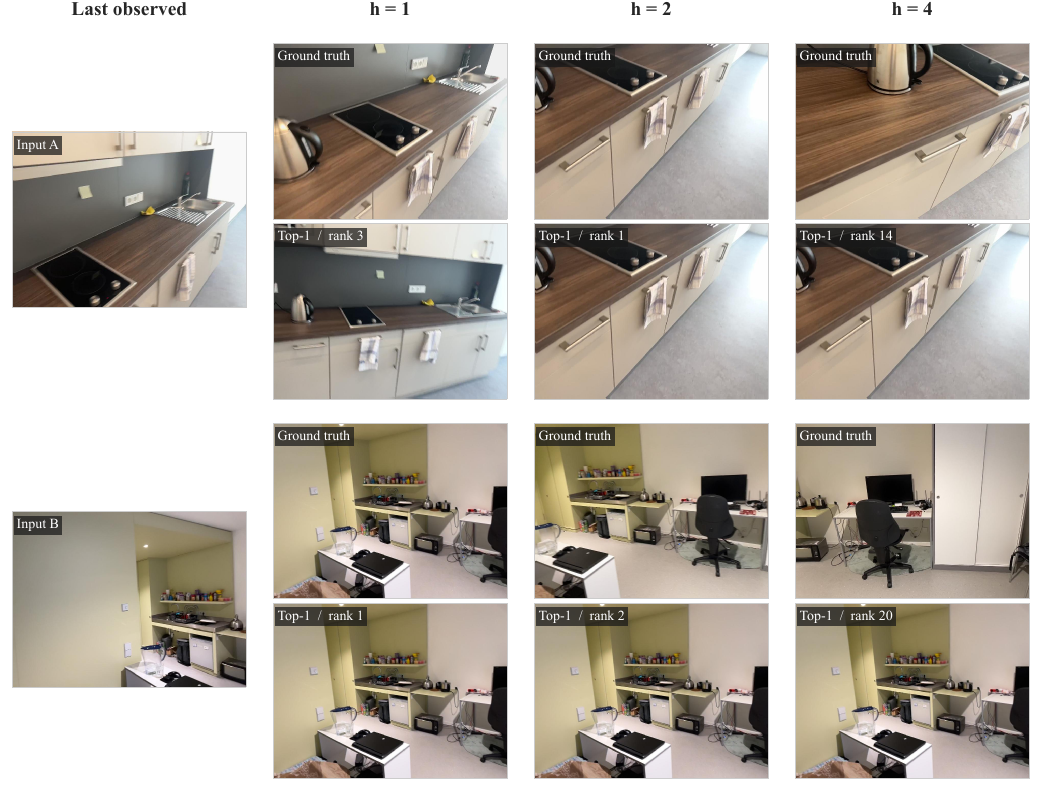}
\caption{Path integration. The Phoenix representation retains distinctions between viewpoints along a camera path, including views of the same room facing in different directions.}
\label{fig:s-probing-path}
\end{figure}

\begin{figure}[!htbp]
\centering
\includegraphics[trim=0 0 42.2bp 0,clip,width=\textwidth,height=.76\textheight,keepaspectratio]{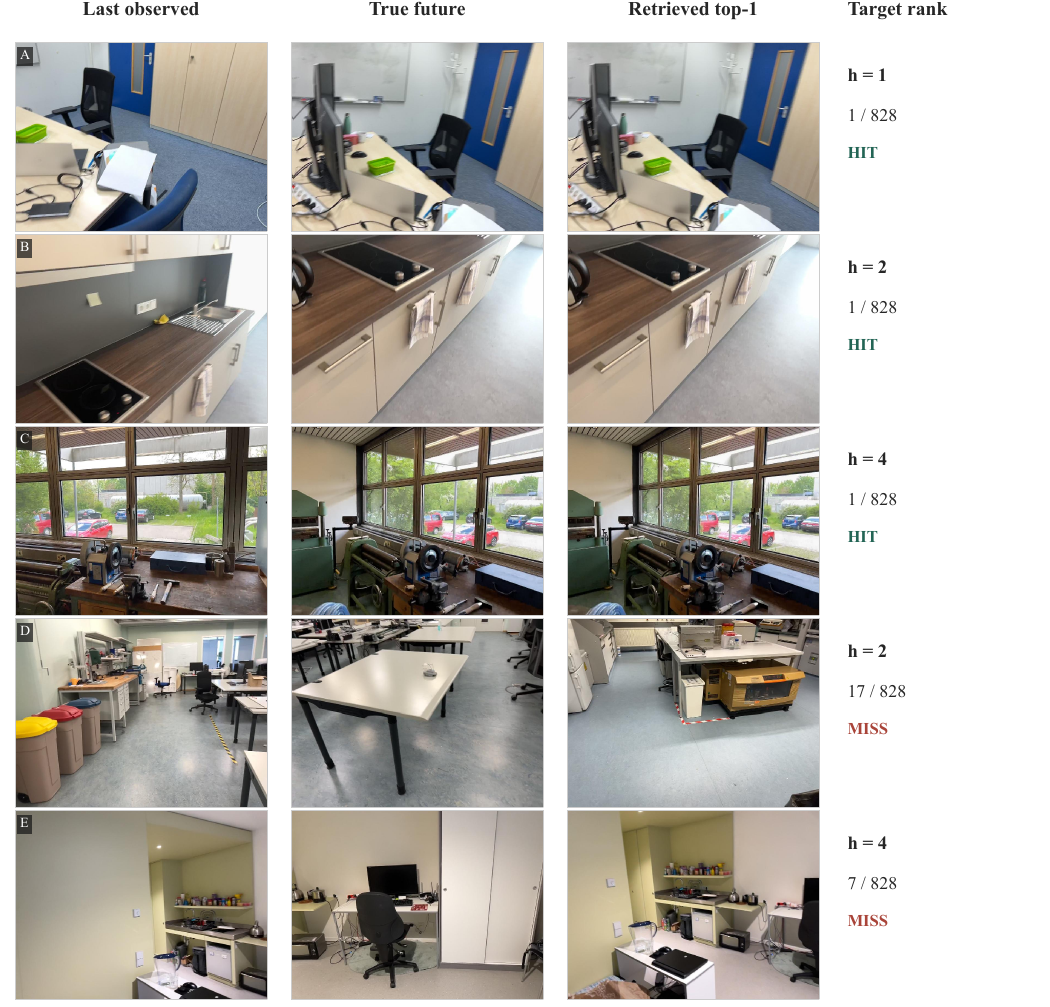}
\caption{Direct future-feature prediction. Given the camera motion, predicted Phoenix features can distinguish the target viewpoint among different views of the same scene.}
\label{fig:s-probing-direct}
\end{figure}

\textit{Scene structure under camera motion.} Future-view prediction requires a finer distinction than overlap recognition: several views may depict the same workspace, but only one corresponds to the specified camera motion. In the kitchen sequence, the initial view looks along the countertop toward the sink. After a downward tilt, both path integration and direct prediction retrieve the exact $h=2$ view of the hob, cabinet fronts, and floor from 828 candidates (Figs.~\ref{fig:s-probing-path}A and~\ref{fig:s-probing-direct}B). In the workshop, direct prediction retrieves the exact $h=4$ view, in which the foreground machinery appears at a different position relative to the windows (Fig.~\ref{fig:s-probing-direct}C). A turn from the kitchenette toward a desk chair and cabinets is missed at the same horizon. Both readouts retrieve the same $h=1$ counter-facing view, with the true $h=4$ target ranked seventh for direct prediction and twentieth for path integration (Figs.~\ref{fig:s-probing-direct}E and~\ref{fig:s-probing-path}B). The successful retrievals show that the frozen Phoenix representation preserves enough viewpoint-specific detail to distinguish the target view from other views of the same scene.

\clearpage\onecolumn

\clearpage\onecolumn

\section{Additional Future-Observation Prediction Visualizations}
\label{sec:s-future-prediction}
\begin{figure}[ht]
\centering
\includegraphics[width=\textwidth,height=0.90\textheight,keepaspectratio]{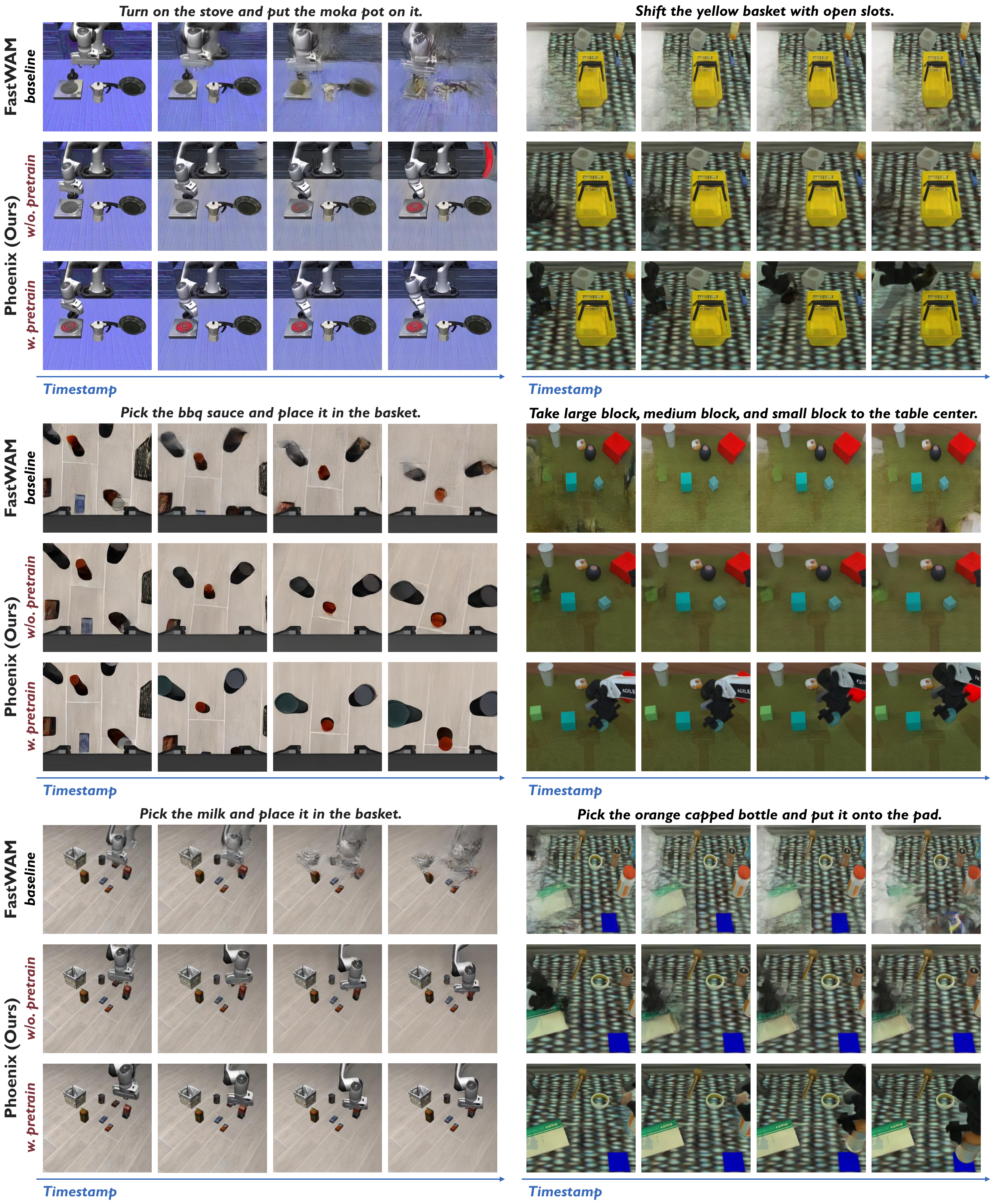}
\caption{Additional future-video predictions on LIBERO-Plus (left) and RoboTwin 2.0 Clean2Random (right).
Pretrained Phoenix produces clearer, more coherent predictions
across six tasks.
}
\label{fig:s-future-prediction}
\end{figure}
\clearpage\twocolumn

\section{Architecture and Encoder Distillation}\label{sec:s-architecture}
\subsection{Phoenix Encoder}
\label{sec:s-encoder}
\label{sec:s-distill}

Phoenix Encoder learns per-frame scene features by distilling
geometric knowledge from VGGT-$\Omega$ and spatiotemporal semantic
knowledge from V-JEPA~2.1-L. Two spatiotemporal adapters connect the
shared image backbone to the frozen teachers during distillation.

\paragraph{Architecture and feature interface.}
We initialize a ViT-L/16 student from the DINOv3 backbone in
VGGT-$\Omega$~\cite{dino,omega}, building on its pretrained visual
semantics. Two spatiotemporal adapters connect the per-frame backbone
to clip-level supervision: a geometry adapter and a JEPA adapter.
Their outputs are projected or predicted in the corresponding teacher
feature spaces during distillation. The adapters use residual output
projections initialized to zero. Table~\ref{tab:s-encoder}
summarizes the architecture.

The retained image backbone processes each frame independently.
For an image $I_t$, it produces
\begin{equation}
F_t=E_{\mathrm{Phoenix}}(I_t)
\in\mathbb{R}^{H_p\times W_p\times D},
\end{equation}
where $H_p,W_p$ index the spatial patch grid and $D$ is the feature
dimension. At the training resolution, each per-frame feature map has
1024 channels and retains a $16\times16$ spatial patch grid. CS-VAE
performs causal aggregation and compression of these features.

\begin{table}[t]
\centering
\footnotesize
\caption{Phoenix Encoder architecture. Adapters, predictor, and
projection heads are used only during distillation.}
\label{tab:s-encoder}
\begin{tabular}{@{}p{.28\linewidth}p{.66\linewidth}@{}}
\toprule
Component & Configuration \\
\midrule
Backbone &
DINOv3 ViT-L/16 initialized from VGGT-$\Omega$ \\
Training input &
$256\times256$, patch size 16 \\
Feature map &
$16\times16\times1024$ per frame \\
Adapters &
Two branches; eight blocks each, width 1024,
16 heads, MLP ratio 4 \\
Auxiliary tokens &
One camera token and 16 register tokens \\
Adapter taps &
Blocks $\{1,3,5,7\}$ \\
JEPA predictor &
Four blocks, width 384, 12 heads;
tubelet convolution $(2,1,1)$ \\
Parameters &
Approximately 300M for the backbone and
320M for the adapter pair \\
\bottomrule
\end{tabular}
\end{table}

\paragraph{Geometric supervision.}
The frozen VGGT-$\Omega$ teacher provides features from layers
$l\in\mathcal J=\{4,11,17,23\}$, using zero-based indices in its
24-layer aggregator. We expand the main-paper alignment objective
over feature groups $s=(l,g)$, where $g=\mathrm{frame}$ and
$g=\mathrm{cross}$ denote per-frame and cross-view patch features.
The set $\mathcal A$ contains four auxiliary groups: per-frame
camera, cross-view camera, per-frame register, and cross-view
register features. With projected student features $\widehat C_{l,g}$,
frozen teacher targets $C_{l,g}$, and mean smooth-$L_1$ distance
$\Delta$, the loss is
\begin{equation}
\begin{aligned}
\mathcal L_{\mathrm{align}}
=\frac{1}{4}\sum_{l\in\mathcal J}\bigl[
&\Delta(\widehat C_{l,\mathrm{frame}},C_{l,\mathrm{frame}})\\
&+0.5\,\Delta(\widehat C_{l,\mathrm{cross}},C_{l,\mathrm{cross}})\\
&+0.05\sum_{g\in\mathcal A}\Delta(\widehat C_{l,g},C_{l,g})\bigr].
\end{aligned}
\end{equation}
Thus, each auxiliary group has weight $0.05$, and the outer factor
$1/4$ averages over teacher layers. In the main-paper notation,
$\lambda_s$ equals the corresponding group weight divided by four.

Projection heads match student and teacher feature dimensions.
For patch groups, both student and teacher features undergo
token-wise layer normalization across channels before alignment;
camera and register groups use the projected student features and
teacher targets directly. Each $\Delta$ averages the elementwise
smooth-$L_1$ loss over the batch, frames, tokens, and channels within
the corresponding group.

\paragraph{Auxiliary depth supervision.}
The frozen VGGT-$\Omega$ depth head produces teacher depth $D_i^{\mathrm T}$
and confidence $c_i$. The same frozen head decodes student depth
$\widehat D_i$ from the projected geometry-adapter features. Its
parameters remain fixed, while gradients through the student path
update the adapter and backbone.

Following the main paper, we use a confidence-weighted log-depth loss:
\begin{equation}
\mathcal L_{\mathrm{depth}}
=\frac{\sum_{i\in\mathcal P}w_i\rho(r_i)}
{\max(\sum_{i\in\mathcal P}w_i,1)}.
\end{equation}
Here, $\mathcal P$ contains all pixels with finite, positive teacher
depth and finite confidence across all clips and frames in the batch.
The numerator and denominator aggregate these pixels jointly, giving
a batch-wide confidence-weighted mean. The denominator is clamped
below by one for numerical stability.

The residual $r_i$ measures the difference in log-depth, and $\rho$
is the Charbonnier penalty:
\begin{equation}
\begin{aligned}
r_i={}&\log\max(\widehat D_i,\epsilon_d)\\
&-\log\max(D_i^{\mathrm T},\epsilon_d),\\
\rho(r)={}&\sqrt{r^2+\epsilon_\rho^2},
\end{aligned}
\end{equation}
where $\epsilon_d=10^{-6}$ clamps depth before taking logarithms
and $\epsilon_\rho=10^{-3}$ smooths the penalty near zero.

For each clip, we compute the confidence quantiles $q_{0.1}$ and
$q_{0.9}$ over all valid pixels across its frames. Each pixel uses
the quantiles of its own clip:
\begin{equation}
\begin{aligned}
\widetilde c_i&=\operatorname{clip}\!\left(
\frac{c_i-q_{0.1}}{q_{0.9}-q_{0.1}+\epsilon_d},0,1\right),\\
w_i&=0.05+0.95\,\widetilde c_i.
\end{aligned}
\end{equation}
Invalid pixels have zero weight. If
$q_{0.9}-q_{0.1}\leq\epsilon_d$, valid pixels in that clip instead
receive unit weights. The positive weight floor retains supervision
from valid pixels with lower teacher confidence.

The complete geometric objective remains
$\mathcal L_{\mathrm{geo}}
=\mathcal L_{\mathrm{align}}+\mathcal L_{\mathrm{depth}}$,
as in the main paper. Teacher depth and confidence are fixed targets;
ground-truth depth annotations are not used.

\paragraph{Spatiotemporal semantic supervision.}
The frozen V-JEPA~2.1-L teacher~\cite{vjepa21} observes the full clip,
while the student receives spatial masks applied consistently across
frames. A masked patch-grid position remains masked throughout the
student clip.
A predictor maps the JEPA adapter outputs into the teacher feature
space. Student predictions and teacher targets undergo token-wise
layer normalization across channels before alignment.

The temporal branch aligns both masked and visible locations:
\begin{equation}
\begin{aligned}
\mathcal L_{\mathrm{tmp}}
={}&\frac{1}{|\mathcal M|}
\sum_{i\in\mathcal M}\Delta(Q_i,Z_i) \\
&+\frac{\lambda_{\mathrm{ctx}}}{|\mathcal V|}
\sum_{i\in\mathcal V}\alpha_i\,\Delta(Q_i,Z_i),
\end{aligned}
\end{equation}
where $\mathcal M$ and $\mathcal V$ contain all masked and visible
spatiotemporal token locations in the batch, and $Q_i$ and $Z_i$
denote the normalized student predictions and teacher targets.
At each location, $\Delta$ averages the elementwise smooth-$L_1$
loss over feature channels. The first term averages this loss over
masked tokens; the second provides distance-weighted visible-context
supervision, scaled by $\lambda_{\mathrm{ctx}}$.

For a visible location $i$, let $d_i$ be the Euclidean distance,
in patch units, to the nearest masked position on the spatial patch
grid of the same clip. We define
\begin{equation}
\begin{aligned}
\widetilde\alpha_i&=(d_i+1)^{-1/2},\\
\alpha_i&=\frac{\widetilde\alpha_i}
{|\mathcal V|^{-1}\sum_{j\in\mathcal V}\widetilde\alpha_j}.
\end{aligned}
\end{equation}
Each spatial weight is shared across temporal locations at the same
patch position within a clip. The normalization is over all visible
spatiotemporal locations in the batch, giving $\alpha_i$ unit mean.
The context term is therefore a distance-weighted mean scaled by
$\lambda_{\mathrm{ctx}}$, with greater weight on nearby visible tokens.

We increase $\lambda_{\mathrm{ctx}}$ linearly from $0$ to $0.5$
over the first 20\% of the distillation schedule, then keep it fixed
at $0.5$. Progress is measured in global optimizer steps, counting
both geometry and temporal updates.

\paragraph{Mask sampling.}
We construct masks on the $16\times16$ spatial patch grid, alternating
between eight small rectangles and two large rectangles across
temporal-branch updates. Each rectangle has a scheduled target area,
an aspect ratio sampled uniformly from $[0.75,1.5]$, and a uniformly
sampled valid position. We merge the rectangles into a mask; its
final coverage is the fraction of masked patches, counting overlaps
only once. We accept the mask if this coverage lies within the
current interval, and otherwise resample the rectangles. Each clip
is sampled independently, and its accepted mask is shared by all
frames.

Over the first 20\% of temporal-branch updates, the target area
\emph{per rectangle} increases linearly from 7\% to 15\% of the grid
for small blocks, and from 25\% to 70\% for large blocks. The allowed
\emph{final coverage} interval changes linearly from $[35\%,55\%]$
to $[50\%,90\%]$. Both settings remain fixed at their final values
thereafter.

\paragraph{Optimization and export.}
Geometric and spatiotemporal semantic objectives alternate at
successive optimizer steps in a $1{:}1$ ratio. Each step updates the
shared backbone and the active branch: the geometry adapter and its
projection heads for geometric supervision, or the JEPA adapter and
predictor for spatiotemporal semantic supervision. Both teachers
remain frozen. Table~\ref{tab:s-opt} summarizes the training settings.

\begin{table}[t]
\centering
\footnotesize
\caption{Phoenix Encoder pretraining settings. Learning rates
are peak values.}
\label{tab:s-opt}
\begin{tabular}{@{}p{.34\linewidth}p{.60\linewidth}@{}}
\toprule
Setting & Value \\
\midrule
Clip length &
Uniformly sampled from $\{4,8,12,16\}$ \\
Temporal sampling &
10 fps; stride 1 or 2 \\
Per-device frame budget &
32 frames per microbatch \\
Optimizer &
AdamW, $(\beta_1,\beta_2)=(0.9,0.95)$,
$\epsilon=10^{-8}$ \\
Weight decay &
0.05 \\
Gradient clipping &
1.0 \\
Backbone learning rate &
$10^{-5}$ \\
Adapter/predictor rate &
$10^{-4}$ \\
Schedule &
5\% warmup followed by cosine decay \\
Precision &
BF16 \\
EMA momentum &
0.9999 \\
Exported checkpoint &
EMA backbone at 110k global steps \\
\bottomrule
\end{tabular}
\end{table}

After each optimizer update, we update the backbone EMA as
$\theta_{\mathrm{EMA}}
\leftarrow m\theta_{\mathrm{EMA}}+(1-m)\theta$,
with $m=0.9999$. Only the EMA backbone is retained for policy training
and deployment. We export the checkpoint at 110k global optimizer
steps, counting both geometric and spatiotemporal semantic updates,
and keep it frozen during CS-VAE training and world--action learning.
Both adapters, the predictor, and the projection heads are discarded,
so the deployed encoder preserves a per-frame feature interface.

\subsection{CS-VAE Architecture}\label{sec:s-csvae}
CS-VAE operates on Phoenix Encoder features with shape $B\times1024\times9\times16\times16$ per view. Three pre-norm spatial Transformer blocks process each frame independently at width 1024. Two causal temporal downsampling stages then reduce nine positions to three, preserving the first position. A final LayerNorm and linear projection produce 96 channels, split into the mean and log-variance of a 48-channel diagonal Gaussian posterior.

Each temporal stage contains a residual bottleneck with width 256 and a causal $(3,1,1)$ convolution, followed by a $(3,1,1)$ downsampling convolution with stride $(2,1,1)$. Temporal residual convolutions use two positions of left padding. Each downsampling stage retains its first position and appends the strided outputs, giving $9\!\rightarrow\!5\!\rightarrow\!3$ positions at frame endpoints $0,4,8$ within the sampled nine-frame sequence. Spatial resolution remains $16\times16$ throughout.

The decoder projects 48 channels back to 1024, applies two temporal upsampling stages, and uses three spatial Transformer blocks followed by LayerNorm. Each upsampling stage uses a causal convolution to expand channels, rearranges subsequent latent positions into pairs of temporal positions, and applies a residual bottleneck. This restores $3\!\rightarrow\!5\!\rightarrow\!9$ positions while retaining the first position.

\begin{table}[t]
\centering\footnotesize
\caption{CS-VAE architecture. Encoder and decoder spatial blocks use the same configuration.}
\label{tab:s-csvae}
\begin{tabular}{@{}p{.43\linewidth}p{.51\linewidth}@{}}\toprule
Setting & Value\\\midrule
Spatial blocks (enc./dec.) & 3 / 3\\
Hidden / FFN width & 1024 / 2048\\
Attention heads / head width & 16 / 64\\
Activation / dropout & GELU / 0\\
Spatial LayerNorm $\epsilon$ & $10^{-12}$\\
Temporal stages (enc./dec.) & 2 / 2\\
Temporal bottleneck width & 256\\
Temporal normalization & Channel RMSNorm, $\epsilon=10^{-6}$\\
Latent channels / spatial grid & 48 / $16\times16$\\
Posterior log-variance range & $[-30,20]$\\
\bottomrule\end{tabular}
\end{table}

\subsection{World--Action Experts and Conditioning}\label{sec:s-wam-architecture}
The video and action experts each contain 30 Transformer blocks. Their residual widths differ, but both project queries, keys, and values to 24 heads of width 128, enabling joint attention in a common 3072-dimensional attention space. Output projections return each stream to its own residual width. Cross-attention, feed-forward layers, and timestep modulation remain specific to each expert (Table~\ref{tab:s-wam-experts}).

\begin{table}[t]
\centering\footnotesize
\caption{World--action expert configurations. Q/K/V width differs from the action expert's residual width.}
\label{tab:s-wam-experts}
\begin{tabular}{@{}lrr@{}}\toprule
Setting & Video expert & Action expert\\\midrule
Transformer blocks & 30 & 30\\
Residual width & 3072 & 1024\\
FFN width & 14336 & 4096\\
Attention heads & 24 & 24\\
Head width & 128 & 128\\
Q/K/V width & 3072 & 3072\\
Input / output dimensions & 96 / 96 & 80 / 80\\
Condition input width & 4096 & 4096\\
Timestep Fourier width & 256 & 256\\
Normalization $\epsilon$ & $10^{-6}$ & $10^{-6}$\\
\bottomrule\end{tabular}
\end{table}

\paragraph{Visual and action tokens.}
Each pretraining sample uses three camera slots. Views are encoded independently. Appearance and scene latents are concatenated along channels, and the views are tiled along width, yielding a $96\times3\times16\times48$ visual tensor. A $(1,2,2)$ patch embedding produces 576 visual tokens: 192 current-observation tokens and 384 future tokens. Missing camera slots are masked. Each of the 32 action steps is linearly projected from 80 to 1024 dimensions, yielding 32 action tokens; a linear output head maps them back to 80 dimensions. Video tokens use three-dimensional rotary positional embeddings; action tokens use one-dimensional rotary embeddings over control steps. Current visual tokens attend only to current visual tokens, future visual tokens attend to all visual tokens, and action tokens attend to all visual and action tokens.

\paragraph{Language, state, and embodiment conditions.}
The frozen UMT5-XXL encoder produces up to 128 language tokens of width 4096; padding is masked. It has 24 layers, 64 attention heads, and FFN width 10240. The current 80-dimensional robot state is mapped by a linear projection and a learned type embedding to one 4096-dimensional token, appended to the language sequence. Each expert projects this context through its own two-layer GELU MLP: $4096\!\rightarrow\!3072\!\rightarrow\!3072$ for video and $4096\!\rightarrow\!1024\!\rightarrow\!1024$ for actions. The action expert additionally prepends eight learned 1024-dimensional tokens selected by the sample's arm ID. The pretraining prompt table has capacity for 28 arm identities; 18 slots are assigned, and the remaining 10 are unused. With $L\leq128$ language tokens, the video context has $L+1$ tokens and the action context $L+9$.

\paragraph{Parameter counts.}
Table~\ref{tab:s-parameter-counts} reports module sizes, including each module's input/output projections. The WAM has 6.023B trainable parameters. Including the frozen Phoenix Encoder backbone, CS-VAE, and Wan VAE gives 7.103B parameters; the frozen UMT5 encoder adds 5.681B when loaded. Pretraining reads cached text embeddings. These totals include the visual decoders and exclude distillation teachers and distillation-only adapters.

\begin{table}[t]
\centering\footnotesize
\caption{Parameter counts for the reported configuration. M denotes $10^6$ parameters; status refers to WAM training.}
\label{tab:s-parameter-counts}
\begin{tabular}{@{}p{.46\linewidth}rl@{}}\toprule
Module & Parameters (M) & Status\\\midrule
Video expert & 5000.97 & Trainable\\
Action expert, including prompts & 1021.26 & Trainable\\
State projection and type embedding & 0.34 & Trainable\\
Phoenix Encoder backbone & 303.23 & Frozen\\
CS-VAE, encoder and decoder & 72.33 & Frozen\\
Wan VAE, encoder and decoder & 704.69 & Frozen\\
UMT5-XXL encoder & 5680.91 & Frozen\\
\bottomrule\end{tabular}
\end{table}

\section{Training Details}\label{sec:s-training}
\subsection{CS-VAE Training}
CS-VAE uses the frozen 110k-step EMA Phoenix Encoder (Sec.~\ref{sec:s-encoder}) and is trained on filtered robot demonstrations, egocentric videos, and UMI recordings.

\paragraph{Feature standardization and reconstruction.}
CS-VAE standardizes each of the 1024 Phoenix feature channels using fixed training-set statistics:
\begin{equation}
\widetilde F_c=(F_c-\mu_c)/\max(\sigma_c,10^{-6}).
\end{equation}
Here, $\mu_c$ and $\sigma_c$ are estimated from 798,720 sampled training windows, aggregating over frames and spatial positions with equal camera weighting within each window. The statistics are stored with the model and reused for all subsequent encoding. Reconstruction losses are evaluated in these standardized coordinates. Decoding to the original feature space applies the inverse transformation.

During CS-VAE training, the encoder samples from its diagonal Gaussian posterior. The objective is
\begin{equation}
\begin{split}
\mathcal L_{\mathrm{CS\text{-}VAE}}={}&
\mathcal L_{\mathrm{MSE}}(\widehat{\widetilde F},\widetilde F)
+\mathcal L_{\mathrm{cos}}(\widehat{\widetilde F},\widetilde F)\\
&+\beta_k D_{\mathrm{KL}}\!\left(q(z\mid\widetilde F)\,\|\,\mathcal N(0,I)\right).
\end{split}
\end{equation}
MSE is averaged over feature elements. The cosine loss is $1-\cos$, computed along channels and averaged over positions. KL divergence is averaged over latent elements. The KL coefficient increases linearly from zero to $10^{-4}$ over the first 30k updates. Downstream WAM training and inference use the posterior mean, with no additional scaling of the CS-VAE latents.

\paragraph{CS-VAE optimization.}
We train for 150k updates with AdamW, global batch size 6144, $(\beta_1,\beta_2)=(0.9,0.99)$, weight decay $10^{-4}$, and gradient clipping at 1.0. During the first 145k updates, the learning rate warms up for 1000 updates to $1.5\times10^{-4}$ and then follows a cosine schedule toward 90\% of this peak. At step 145k, it is reset to $5\times10^{-5}$ and decays to $3\times10^{-5}$ over the final 5k updates. Training uses BF16 autocast with FP32 parameters and feature statistics. The 150k-step checkpoint is frozen for WAM learning.

\subsection{World--Action Pretraining and Post-training}
\paragraph{Initialization and trainable modules.}
The video expert is initialized from Wan2.2-TI2V-5B. Its 48-channel input/output projections are expanded to 96 channels: the appearance-channel weights are copied and the added scene-channel weights are zero-initialized. The action expert is randomly initialized. Phoenix Encoder, CS-VAE, and Wan VAE remain frozen; the two experts, state projection, and soft prompts are optimized jointly. Text embeddings are precomputed with the frozen UMT5 encoder.

\paragraph{Sampling and optimization.}
WAM pretraining uses robot demonstrations. A sample spans 33 consecutive control timestamps, from which we select nine frames at indices $\{0,4,\ldots,32\}$ for each camera and 32 action targets. Each view has resolution $256\times256$. AdamW uses a global batch size of 4480, $(\beta_1,\beta_2)=(0.9,0.95)$, weight decay 0.01, and gradient clipping at 1.0. The full-data schedule has 256,013 updates, including 10k warmup updates to $10^{-4}$ and cosine decay to $10^{-6}$. Training uses BF16 and DeepSpeed ZeRO-1, with 20 samples per GPU across 224 GPUs. The scaling experiments train for one epoch at each data scale.

\paragraph{Post-training.}
Post-training starts from the final checkpoint of the corresponding one-epoch WAM pretraining run and retains the same objective and frozen/trainable partition. We use a learning rate of $5\times10^{-5}$ and a global batch size of 256, adapting the model separately to each benchmark. Table~\ref{tab:s-posttrain-protocol} lists the post-training updates and action execution intervals.

\begin{table}[t]
\centering\footnotesize
\caption{Post-training and action execution. Execution steps are the number of predicted actions applied before replanning; each prediction contains 32 actions.}
\label{tab:s-posttrain-protocol}
\begin{tabular}{@{}p{.52\linewidth}rr@{}}\toprule
Benchmark & Updates & Execution steps\\\midrule
LIBERO-Plus / LIBERO-Pro & 25k & 10\\
RoboTwin 2.0 C2R & 25k & 32\\
RoboCasa-GR1 & 30k & 12\\
Bench2Dex & 20k & 8\\
RoboDojo-RealWorld & 30k & 32\\
\bottomrule\end{tabular}
\end{table}

\paragraph{Flow-matching objective.}
We use flow matching~\cite{flow}, sampling video and action noise levels independently. For $u\sim\mathcal U(0,1)$, the shifted noise level and perturbed target are
\begin{equation}
\sigma=\frac{s u}{1+(s-1)u},\qquad
x_\sigma=(1-\sigma)x+\sigma\epsilon,
\end{equation}
where $\epsilon\sim\mathcal N(0,I)$, $s=6$ for video, and $s=1$ for actions. Each expert predicts the flow target $\epsilon-x$ with timestep embedding input $1000\sigma$. The current visual latent remains clean and is excluded from prediction loss. The joint objective is
\begin{equation}
\mathcal L_{\mathrm{WAM}}=
0.5\mathcal L_{\mathrm{appearance}}
+0.5\mathcal L_{\mathrm{feature}}
+\mathcal L_{\mathrm{action}}.
\end{equation}
Each term first averages the squared flow error over valid elements within each sample, applies that sample's timestep weight, and then averages over the batch. The scheduler weight is proportional to $e^{-2(\sigma-1/2)^2}-m_s$, where $m_s$ is the minimum on its 1000-point shifted time grid; weights are normalized to unit mean on that grid. Invalid camera slots, padded targets, and unavailable control dimensions are masked out.

\subsection{Inference}
At each decision step, the frozen visual modules encode the current observations, and the state and language provide the conditioning context described in Sec.~\ref{sec:s-wam-architecture}. Future visual latents and the 32-step action chunk are initialized from Gaussian noise and jointly denoised with 10 Euler steps, using the same video/action shifts as training. The current visual latents stay fixed. We execute the action prefix specified in Table~\ref{tab:s-posttrain-protocol} before replanning. Predicted appearance latents can be decoded to visualize scene evolution; action outputs are denormalized and mapped to the robot's control interface.

\section{Pretraining Data and Processing}\label{sec:s-data}
\subsection{Data Composition}
\begin{figure*}[t]
\centering
\includegraphics[width=\textwidth]{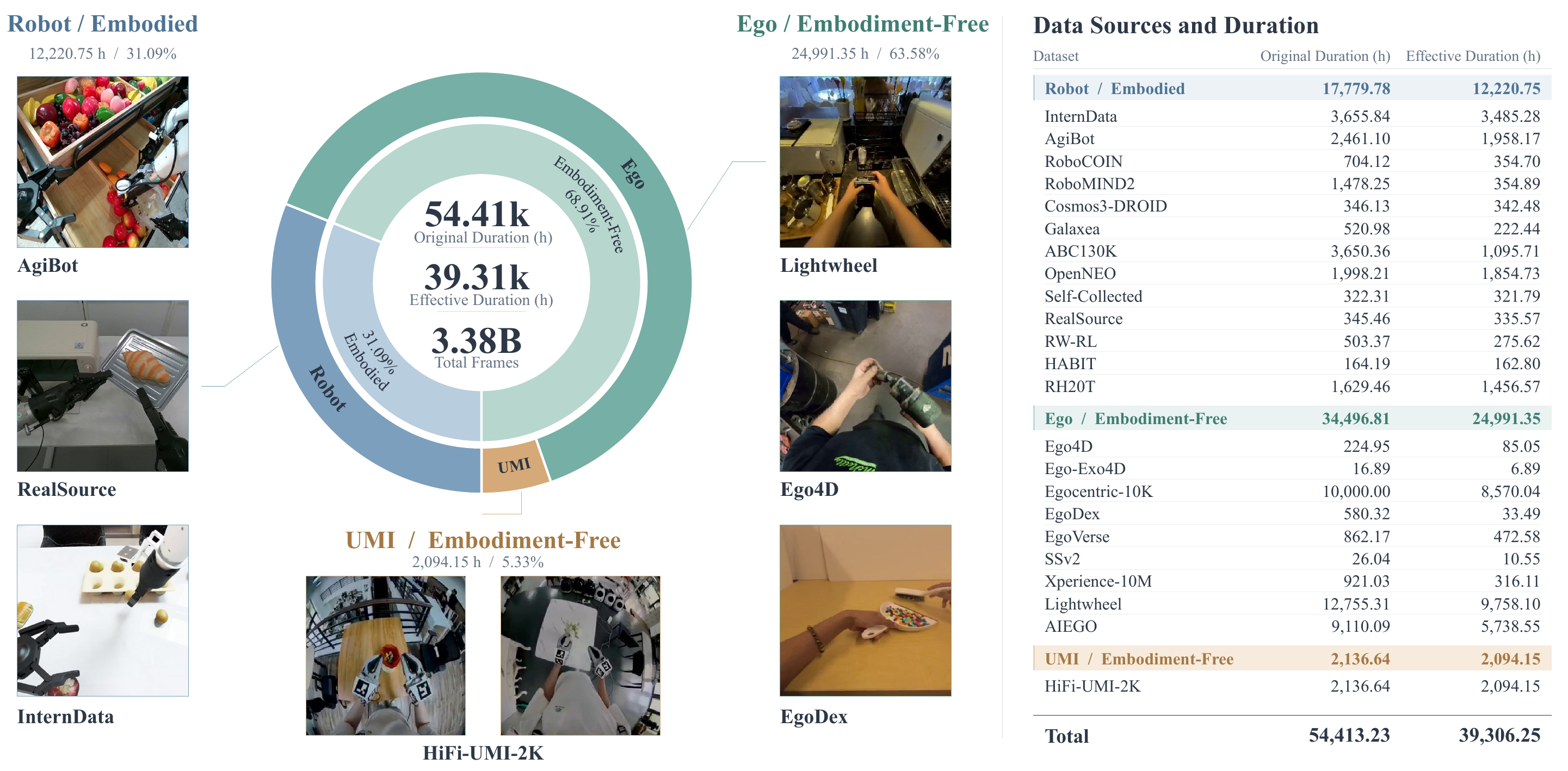}
\caption{\textbf{Composition of the pretraining data pool.} Original and retained durations are reported for each source. Italicized dataset names indicate non-public datasets.}
\label{fig:s-data-pool}
\end{figure*}

Our pretraining pool comprises robot demonstrations, egocentric videos, and Universal Manipulation Interface (UMI) recordings, totaling 54,413 hours of raw data and approximately 3.38 billion frames. Filtering retains 39,306 hours, of which robot demonstrations, egocentric videos, and UMI recordings account for 31.09\%, 63.58\%, and 5.33\% by duration, respectively. Figure~\ref{fig:s-data-pool} presents the sources, representative observations, and original and retained durations.

Robot data combine public datasets and in-house demonstrations spanning different arms, end effectors, camera configurations, and manipulation tasks. They provide visual, state, and action supervision grounded in robot control. Egocentric videos and UMI recordings contribute manipulation sequences from additional scenes, broadening the coverage of visual representation learning.

The three pretraining stages use different portions of this pool. Encoder distillation and CS-VAE training use all filtered data to learn scene representations and their spatiotemporal compression, respectively. World--action pretraining uses only robot demonstrations to jointly predict future visual latents and action chunks.

\subsection{Robot Data Processing Pipeline}
\begin{figure*}[t]
\centering
\includegraphics[width=\textwidth]{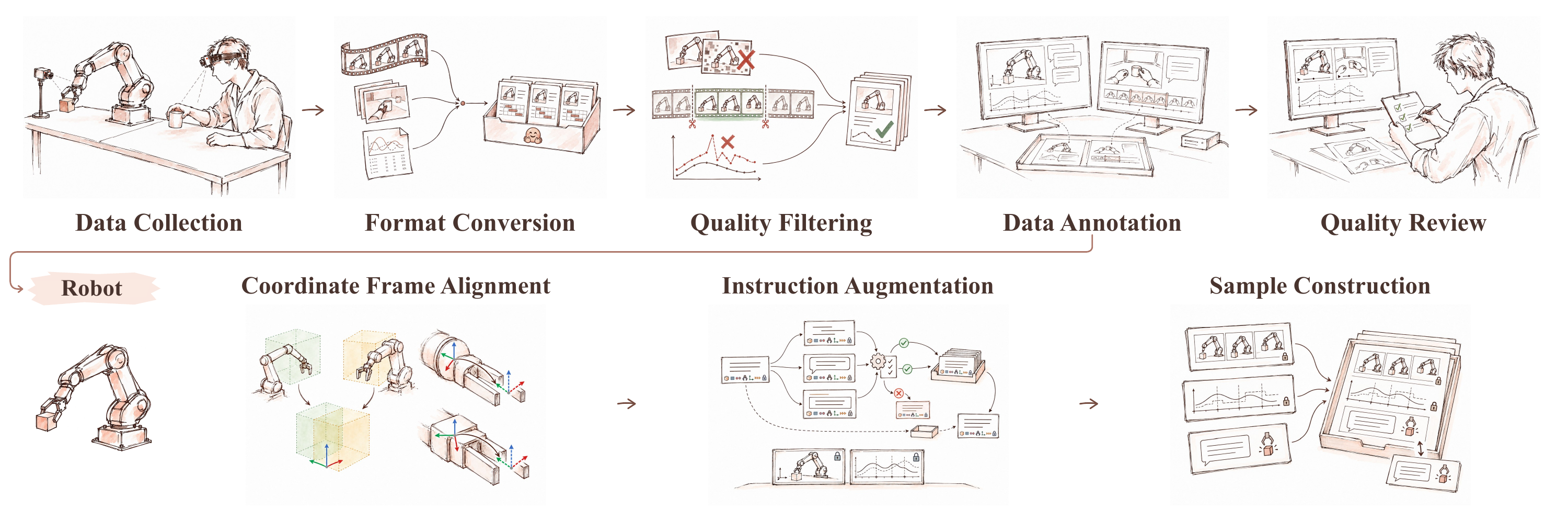}
\caption{\textbf{Robot data processing pipeline.} Collected demonstrations undergo format conversion, quality filtering, annotation, and review. Coordinate-frame alignment and instruction augmentation prepare synchronized observations, states, actions, and instructions for sample construction.}
\label{fig:s-robot-data-pipeline}
\end{figure*}

Our robot data curation follows practices in Qwen-RobotManip~\cite{qwen} and related work, emphasizing consistency among visual observations, robot states, actions, and task instructions. Sources differ not only in file format but also in timestamps, coordinate conventions, and the meanings of control fields. We therefore first standardize the format and check temporal correspondence, then inspect visual and motion quality, trim redundant episode boundaries, and align control semantics. Instruction augmentation, sample construction, and manual review complete the process (Fig.~\ref{fig:s-robot-data-pipeline}).

\paragraph{Format conversion and temporal synchronization.}
We convert videos, states, actions, and task information to LeRobot V3, organize them by complete trajectory, and specify joint order, position units, rotation representations, and gripper semantics. Measured states and action commands are treated separately: the former describe actual robot motion, whereas the latter specify requested control and may differ because of contact, control latency, or motion constraints. We associate video presentation timestamps with state acquisition times and check matching errors. Videos are fully decoded after conversion to verify frame counts, state rows, and task boundaries. For sources requiring one-to-one frame correspondence, we verify a complete one-to-one mapping between video frames and state records.

\paragraph{Visual quality checks.}
After decoding and synchronization checks, we assess whether videos retain recognizable manipulation details. We sample frames uniformly, convert them to grayscale, and resize them to a common scale. Sharpness is measured by Laplacian-response energy normalized by gradient energy, reducing sensitivity to image contrast. The default configuration samples eight frames per clip and resizes the shorter side to 256 pixels. A clip is flagged as a blur candidate when its median normalized sharpness is below 0.18 and fewer than 50\% of sampled frames reach this threshold; candidates are reviewed against the video.

\paragraph{Motion checks and static boundary trimming.}
We sample short-interval frame pairs at different positions in a video, estimate pixel displacement using optical flow, and normalize it by image size and the frame interval. Clip-level statistics capture both typical motion and short bursts to identify prolonged inactivity or abrupt changes. Optical flow and sharpness provide complementary checks: the former measures image motion, while the latter assesses whether recognizable details remain visible during that motion.

For redundant stationary segments at episode boundaries, we further use measured positions, rotations, joints, and gripper states to determine the retained interval. Detection proceeds inward from each end, constraining both adjacent-state changes and cumulative changes relative to that end's initial state. The candidate stationary interval expands only while every monitored channel satisfies its stationary criterion, and stops when any channel exhibits valid motion. The cumulative constraint preserves slow operations whose per-step displacement is small but accumulates into meaningful movement. Rotation differences are measured as relative rotation angles on SO(3), avoiding artifacts from Euler-angle wrapping or quaternion sign flips.

Trimming applies only to contiguous stationary prefixes and suffixes; pauses and contact phases inside a task are retained. We preserve four state points as boundary context and keep at least 33 state points in trajectories that were originally long enough. Missing or invalid required states stop further expansion of a stationary interval, while entirely stationary trajectories are flagged for review. Videos, states, actions, and annotations are sliced together, with indices and statistics updated to keep subsequent sampling within a consistent valid time range.

\paragraph{Coordinate and control-semantic alignment.}
After temporal synchronization and boundary trimming, we check and standardize each field's physical meaning, including position units, rotation conventions, reference and end-effector frames, left/right arm assignment, gripper direction and value semantics, and absolute or incremental control semantics. These fields are then mapped to the unified state--action interface; Sec.~\ref{sec:s-interface} specifies the coordinate transforms, action definitions, and slot mapping.

\paragraph{Instruction processing and augmentation.}
Once visual and action supervision is established, we diversify language through instruction rewriting. Each eligible instruction receives three semantic paraphrases, two conversational variants, and three context-framing variants. Rewriting is constrained to preserve objects and attributes, quantities, arm assignments, spatial relations, action order, and task constraints; context framing adds no new scene facts or manipulation requirements. Augmentation is applied to complete instructions with unambiguous, internally consistent task content. Outputs are checked for candidate counts, nonempty text, and duplicates, with the original instruction retained when no usable augmentation is available. All variants remain associated with the original task and leave video and action supervision unchanged.

\paragraph{Training sample construction.}
We construct continuous windows from processed trajectories, pairing a task instruction with multiview observations, the current robot state, and future actions. Each window contains 33 consecutive state timestamps to construct a 32-step action chunk, with nine RGB frames sampled per view. The complete window, including the endpoint state needed for the final action, must lie within one continuous task interval of a single trajectory. Repeated occurrences of the same task are not joined across intervening tasks. Intervals that are too short are not padded with repeated states or frames to create training windows.

\paragraph{Manual semantic verification.}
Manual review uses synchronized videos, end-effector poses, and motion traces to verify motion direction, arm assignment, end-effector orientation, gripper semantics, temporal alignment, and consistency between instructions and observed actions. Review samples include ordinary retained demonstrations as well as potentially problematic crop boundaries, motion discontinuities, and complex instructions.

\subsection{Unified Action Space}\label{sec:s-interface}
We map source-specific robot states and controls to fixed semantic slots in an 80-dimensional interface after coordinate and control-semantic alignment.

\begin{figure*}[t]
\centering
\includegraphics[width=\textwidth]{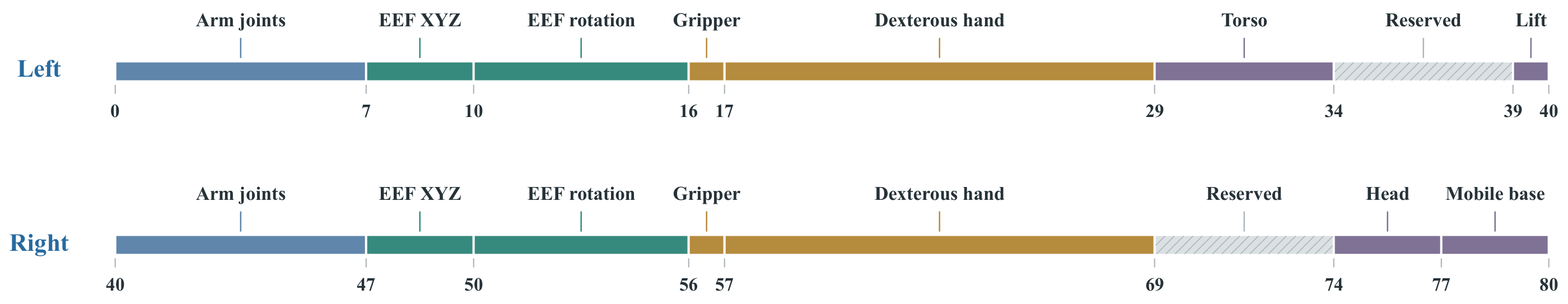}
\caption{\textbf{Unified 80-dimensional state--action interface.} Slot boundaries use zero-based indexing. The two rows show the left-arm and right-arm portions together with body-control and reserved slots. Single-arm data use the right-arm slots. End-effector states use all nine pose dimensions, whereas end-effector actions use the first six and mask the remaining three. Missing and reserved dimensions are masked out of action supervision.}
\label{fig:s-action-space}
\end{figure*}

\paragraph{Coordinate-frame alignment.}
For sources containing end-effector poses, we convert positions to meters and interpret rotations using source-specific conventions before aligning reference-frame and local end-effector axes separately. Let $p$ and $R$ be the original end-effector position and rotation, $A$ the reference-frame rotation, and $B$ the local end-effector axis rotation. The aligned pose is
\begin{equation}
\widetilde p=Ap,\qquad \widetilde R=ARB.
\end{equation}
Here, $A$ maps source axes to the convention $+X$ forward, $+Y$ left, and $+Z$ up, while $B$ is specified for the robot configuration and arm side. This transformation preserves the source's selected end-effector position reference point.

\paragraph{State and action representations.}
States describe the current robot configuration, including absolute joint positions, end-effector poses, and gripper or dexterous-hand states. An end-effector state uses a three-dimensional position and a continuous six-dimensional rotation representation, totaling nine dimensions. Joint actions specify absolute target joint positions. End-effector actions use a three-dimensional translation increment and a three-dimensional rotation vector:
\begin{equation}
\begin{aligned}
\Delta p &= \widetilde p_{\mathrm{target}}-\widetilde p_t,\\
\Delta r &= \operatorname{Log}\!\left(\widetilde R_t^{\top}\widetilde R_{\mathrm{target}}\right)^\vee.
\end{aligned}
\end{equation}
The translation increment is expressed in the aligned reference frame, and the rotation vector expresses the relative rotation in the current end-effector frame. Gripper actions use absolute scalar controls.

\paragraph{Fixed slots and validity masks.}
The interface allocates fixed slots to left/right arm joints, end-effector positions and rotations, grippers, dexterous hands, torso, lift, head, and mobile-base controls (Fig.~\ref{fig:s-action-space}). Single-arm data use the right-arm slots. End-effector states occupy all nine pose dimensions, whereas end-effector actions use only the first six, with the remaining dimensions masked. Missing, reserved, or invalid dimensions are zero-filled and excluded from action loss.

\paragraph{Normalization.}
After field mapping and coordinate alignment, we group data by source and robot configuration and compute separate training-set statistics for states and actions. Ordinary continuous dimensions use normalization based on the 1st and 99th percentiles to reduce scale differences across control quantities; the continuous six-dimensional rotation representation in the state is retained directly. Statistics use only valid training observations, excluding validation data and structural padding. Inference reuses the training statistics to convert model outputs back to the corresponding robot controls.

\paragraph{Embodiment conditioning.}
Embodiment-specific soft prompts condition the shared policy on robot configuration and control characteristics. Inspired by the learnable soft prompts in X-VLA~\cite{xvla}, each arm ID selects eight learnable tokens, forming an embodiment condition $P_e\in\mathbb{R}^{8\times d}$. These tokens join language and robot state in action cross-attention and are jointly trained with the policy.


\clearpage
\phantomsection
\pdfbookmark[0]{References}{shared-references}
{\small\setlength{\bibsep}{0pt}
\bibliographystyle{ieeenat_fullname}
\bibliography{paper/references,paper/table_refs,paper/added_refs,paper/spatial_probing_refs,paper/supplement_refs,paper/supplement_official_refs}}
\end{document}